\documentclass{article}

\PassOptionsToPackage{numbers, sort}{natbib}

\usepackage[preprint]{neurips_2021}

\usepackage[utf8]{inputenc} 
\usepackage[T1]{fontenc}    
\usepackage[colorlinks]{hyperref}     
\usepackage{url}            
\usepackage{amsfonts}       
\usepackage{nicefrac}       
\usepackage{microtype}      
\usepackage{xcolor}         

\usepackage{xcolor}
\usepackage{caption}
\usepackage{amsmath}
\usepackage{amssymb}
\usepackage{multirow}
\usepackage{graphicx}
\usepackage{booktabs}
\usepackage[export]{adjustbox}

\newcommand{\op}[1]{\operatorname{#1}}
\newcommand{\mbf}[1]{\mathbf{#1}}

\usepackage{algorithm}
\usepackage{listings}
\usepackage{wrapfig,booktabs}
\usepackage{multirow}

\usepackage{caption}
\usepackage{subcaption}
\usepackage{mathtools}

\title{SAINT: Improved Neural Networks for Tabular Data via Row Attention and Contrastive Pre-Training}

\author{%
   Gowthami Somepalli \\
   Department of Computer Science \\
   University of Maryland, College Park \\
   \texttt{gowthami@umd.edu}
   \And
   Micah Goldblum \\
   Department of Computer Science \\
   University of Maryland, College Park \\
   \texttt{goldblum@umd.edu}
   \And
   Avi Schwarzschild \\
   Department of Mathematics \\
   University of Maryland, College Park \\
   \texttt{avi1@umd.edu}
   \And
   C. Bayan Bruss \\
   Capital One \\
   Center for Machine Learning\\
   \texttt{bayan.bruss@capitalone.com}
    \And
   Tom Goldstein \\
   Department of Computer Science \\
   University of Maryland, College Park \\
   \texttt{tomg@umd.edu}
}

\begin{document}

\maketitle

\begin{abstract}
Tabular data underpins numerous high-impact applications of machine learning from fraud detection to genomics and healthcare.  Classical approaches to solving tabular problems, such as gradient boosting and random forests, are widely used by practitioners.  However, recent deep learning methods have achieved a degree of performance competitive with popular techniques.  We devise a hybrid deep learning approach to solving tabular data problems.  Our method, SAINT, performs attention over both rows and columns, and it includes an enhanced embedding method.  We also study a new contrastive self-supervised pre-training method for use when labels are scarce.  SAINT consistently improves performance over previous deep learning methods, and it even outperforms gradient boosting methods, including XGBoost, CatBoost, and LightGBM, on average over a variety of benchmark tasks.
\end{abstract}

\section{Introduction}
\label{sec:introduction}
While machine learning for image and language processing has seen major advances over the last decade, many critical industries, including financial services, health care, and logistics, rely heavily on data in structured table format. 
Tabular data is unique in several ways that have prevented it from benefiting from the impressive success of deep learning in vision and language. First, tabular data often contain heterogeneous features that represent a mixture of continuous, categorical, and ordinal values, and these values can be independent or correlated. Second, there is no inherent positional information in tabular data, meaning that the order of columns is arbitrary. This differs from text, where tokens are always discrete, and ordering impacts semantic meaning.  It also differs from images, where pixels are typically continuous, and nearby pixels are correlated. Tabular models must handle features from multiple discrete and continuous distributions, and they must discover correlations without relying on the positional information. Sufficiently powerful deep learning systems for tabular data have the potential to improve performance beyond what is achieved by classical methods, like linear classifiers and random forests.  Furthermore, without performant deep learning models for tabular data, we lack the ability to exploit compositionality, end-to-end multi-task models, fusion with multiple modalities (e.g. image and text), and representation learning. 

We introduce SAINT, the Self-Attention and Intersample Attention Transformer, a specialized architecture for learning with tabular data. SAINT leverages several mechanisms to overcome the difficulties of training on tabular data. SAINT projects all features -- categorical and continuous -- into a combined dense vector space. These projected values are passed as tokens into a transformer encoder which uses attention in the following two ways. First, there is ``self-attention,'' which attends to individual features within each data sample. Second, we propose a novel ``intersample attention,'' which enhances the classification of a row (i.e., a data sample) by relating it to other rows in the table.  Intersample attention is akin to a nearest-neighbor classification, where the distance metric is learned end-to-end rather than fixed. In addition to this hybrid attention mechanism, we also leverage self-supervised contrastive pre-training to boost performance for semi-supervised problems.

We provide comparisons of SAINT to a wide variety of deep tabular architectures and commonly used tree-based methods using a diverse battery of datasets. We observe that SAINT, on average, outperforms all other methods on supervised and semi-supervised tasks.  More importantly, SAINT often out-performs boosted trees (including  XGBoost \cite{chen2016xgboost}, CatBoost \cite{dorogush2018catboost}, and LightGBM \cite{ke2017lightgbm}), which have long been an industry favorite for complex tabular datasets.
Finally, we visualize the attention matrices produced by our models to gain insights into how they behave.


\section{Related Work}
\label{sec:related_work}
\textbf{Classical Models}
The most widely adopted approaches for supervised and semi-supervised learning on tabular datasets eschew neural models due to their black-box nature and high compute requirements. When one has reasonable expectations of linear relationships, a variety of modeling approaches are available \cite{wright1995logistic,weisberg2005applied,starkweather2011multinomial,mcculloch2005generalized}. In more complex settings, non-parametric tree-based models are used. Commonly used tools such as XGBoost~\cite{chen2016xgboost}, CatBoost~\cite{dorogush2018catboost}, and LightGBM~\cite{ke2017lightgbm} provide several benefits such as interpretability, the ability to handle a variety of feature types including null values, as well as performance in both high and low data regimes.


\textbf{Deep Tabular Models} While classical methods are still the industry favorite, some recent work brings deep learning to the tabular domain. For example, TabNet \cite{arik2019TabNet} uses neural networks to mimic decision trees by placing importance on only a few features at each layer. The attention layers in that model do not use the regular dot-product self-attention common in transformer-based models, rather there is a type of sparse layer that allows only certain features to pass through. \citet{yoon2020vime} propose VIME, which employs MLPs in a technique for pre-training based on denoising. TABERT \cite{yin2020tabert}, a more elaborate neural approach inspired by the large language transformer model BERT \cite{devlin2018bert}, is trained on semi-structured test data to perform language-specific tasks. Several other studies utilize tabular data, but their problem settings are outside of our scope \cite{pathak2016context,chen2019learning,padhi2021tabular,shavitt2018regularization, katzir2021netdnf}.

Transformer models for more general tabular data include
TabTransformer~\cite{huang2020tabtransformer}, which uses a transformer encoder to learn contextual embeddings \emph{only} on categorical features. The continuous features are concatenated to the embedded features and fed to an MLP. The main issue with this model is that continuous data do not go through the self-attention block. That means any information about correlations between categorical and continuous features is lost. In our model, we address that issue by projecting continuous features and categorical features to the higher dimensional embedding space and passing them both through the transformer blocks. In addition, we propose a new type of attention to explicitly allow data points to attend to each other to get better representations.

\textbf{Axial Attention} \citet{ho2019axial} are the first to propose row and column attention in the context of localized attention in 2-dimensional inputs (like images) in their Axial Transformer. This is where for a given pixel, the attention is computed only on the pixels that are on the same row and column, rather than using all the pixels in the image. The MSA Transformer~\cite{rao2021msa} extends this work to protein sequences and applies both column and row attention across similar rows (tied row attention). TABBIE~\cite{iida2021tabbie} is an adaptation of axial attention that applies self-attention to rows and columns separately, then averages the representations and passes them as input to the next layer. In all these works, different features from the same data point communicate with each other and with the same feature from a whole batch of data. Our approach, intersample attention, is hierarchical in nature; first features of a given data point interact with each other, then data points interact with each other using entire rows/samples. 

In a similar vein, Graph Attention Networks~(GAT)~\cite{velivckovic2017graph} seek to compute attention over neighbors on a graph, thereby learning which neighbor's information is most relevant to a given node’s prediction. One way to view our intersample attention is as a GAT on a complete graph where all tabular rows are connected to all other rows. \citet{yang2016hierarchical} explore hierarchical attention for the task of document classification where attention is computed between words in a given sentence and then between the sentences, but they did not attempt to compute the attention between entire documents themselves.

\textbf{Self-Supervised Learning} Self-supervision via a `pretext task' on unlabeled data coupled with finetuning on labeled data is widely used for improving model performance in language and computer vision.  
Some of the tasks previously used for self-supervision on tabular data include masking, denoising, and replaced token detection. Masking (or Masked Language Modeling(MLM)) is when individual features are masked and the model's objective is to impute their value~\cite{pathak2016context,arik2019TabNet,huang2020tabtransformer}. Denoising injects various types of noise into the data, and the objective there is to recover the original values~\cite{vincent2008extracting,yoon2020vime}. Replaced token detection (RTD) inserts random values into a given feature vector and seeks to detect the location of these replacements~\cite{huang2020tabtransformer,iida2021tabbie}. Contrastive pre-training, where the distance between two views of the same point is minimized while maximizing the distance between two different points~\cite{chen2020simple,he2020momentum,grill2020bootstrap}, is another pretext task that applies to tabular data.  In this paper, to the best of our knowledge, we are the first to adopt contrastive learning for tabular data.  We couple this strategy with denoising to perform pre-training on a plethora of datasets with varied volumes of labeled data, and we show that our method outperforms traditional boosting methods.

\section{Self-Attention and Intersample Attention Transformer (SAINT)}
\label{sec:saint}

\begin{figure}[h]
  \centering
  \includegraphics[width=\textwidth]{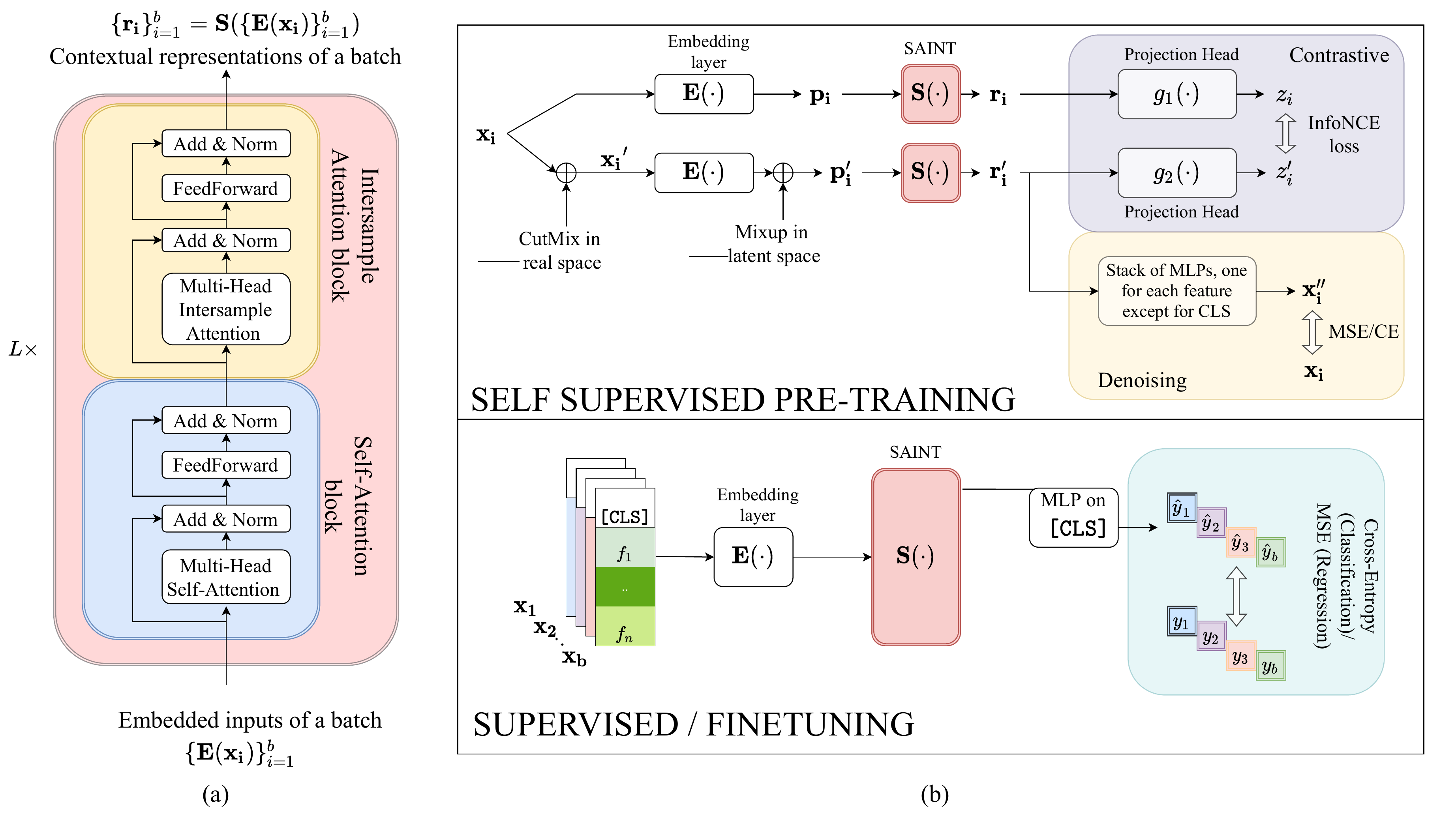}
  \captionsetup{font=small}
  \caption{The SAINT architecture, including pre-training and training pipelines. (a) Inspired by \citep{vaswani2017attention}, we use $L$ layers with 2 attention blocks each, one self-attention block, and one of our novel intersample attention blocks that computes attention across samples (see Section~\ref{subsec:ISA}). (b) For pre-training, we minimize contrastive and denoising losses between a given data point and its views generated by CutMix and mixup (Section~\ref{sec:pretraining}). During finetuning/regular training, data passes through an embedding layer and then the  SAINT model. We take the contextual embeddings from SAINT and pass only the embedding correspond to the CLS token through an MLP to obtain the final prediction.}
  \label{fig:saint_arch_training}
\end{figure}

In this section, we introduce our model, Self-Attention and Intersample Attention Transformer~(SAINT), and explain in detail its various components.

Suppose $\mathcal{D} = \{\mathbf{x_i},y_i\}_{i=1}^m$ is a tabular dataset with $m$ points, where each ${x_i}$ is an $n$-dimensional feature vector, and ${y_i}$ is a label or target variable. Similar to BERT~\citep{devlin2018bert}, we append a \verb|[CLS]| token with a learned embedding to each data sample. Let $\mathbf{x_i} = [\verb|[CLS]|, f_i^{\{1\}}, f_i^{\{2\}},  ..,f_i^{\{n\}}]$ be a single data-point with categorical or continuous features $f_i^{\{j\}}$, and let $\mathbf{E}$ be the embedding layer that embeds each feature into a $d$-dimensional space.  Note that $\mathbf{E}$ may use different embedding functions for different features. For a given $\mathbf{x_i} \in \mathbb{R}^{(n+1)}$, we get $\mathbf{E}(\mathbf{x_i}) \in \mathbb{R}^{(n+1) \times d}$.

\textbf{Encoding the Data} In language models, all tokens are embedded using the same procedure. However, in the tabular domain, different features can come from distinct distributions, necessitating a heterogeneous embedding approach.  Note that tabular data can contain multiple categorical features which may use the same set of tokens. Unless it is known that common tokens possess identical relationships within multiple columns, it is important to embed these columns independently.  Unlike the embedding of TabTransformer\cite{huang2020tabtransformer}, which uses attention to embed only categorical features, we propose also projecting continuous features into a $d-$dimensional space before passing their embedding through the transformer encoder. To this end, we use a separate single fully-connected layer with a ReLU nonlinearity for each continuous feature, thus projecting the $1-$dimensional input into $d-$dimensional space. With this simple trick alone, we significantly improve the performance of the TabTransformer model as discussed in Section \ref{subsec:main_results}. An additional discussion concerning positional encodings can be found in Appendix \ref{appendix_sec:complete_training_deets}.
 

\subsection{Architecture}

SAINT is inspired by the transformer encoder of \citet{vaswani2017attention}, designed for natural language, where the model takes in a sequence of feature embeddings and outputs contextual representations of the same dimension. A graphical overview of SAINT is presented in Figure \ref{fig:saint_arch_training}(a).

SAINT is composed of a stack of $L$ identical stages. Each stage consists of one self-attention transformer block and one intersample attention transformer block. The self-attention transformer block is identical to the encoder from \cite{vaswani2017attention}. It has a multi-head self-attention layer~(MSA) (with $h$ heads), followed by two fully-connected feed-forward (FF) layers with a GELU non-linearity~\cite{hendrycks2016gaussian}. Each layer has a skip connection~\cite{he2016deep} and layer normalization (LN)~\cite{ba2016layer}. The intersample attention transformer block is similar to the self-attention transformer block, except that the self-attention layer is replaced by an intersample attention layer~(MISA). The details of the intersample attention layer are presented in the following subsection.

The SAINT pipeline, with a single stage ($L=1$) and a batch of $b$ inputs, is described by the following equations. We denote multi-head self-attention by MSA, multi-head intersample attention by MISA, feed-forward layers by FF, and layer norm by LN: 
\begin{align}
    \mbf{z_i^{(1)}} &= \op{LN}(\op{MSA}(\mathbf{E}(\mathbf{x_i}))) + \mathbf{E}(\mathbf{x_i}) & \mbf{z_i^{(2)}} &= \op{LN}(\op{FF_1}(\mbf{z_i^{(1)}})) +\mbf{z_i^{(1)}}
    \label{eq:sa_block_apply} \\
    \mbf{z_i^{(3)}} &= \op{LN}(\op{MISA}(\{\mbf{z_i^{(2)}}\}_{i=1}^b)) +\mbf{z_i^{(2)}} &  \mathbf{r_i}&= \op{LN}(\op{FF_2}(\mbf{z_i^{(3)}})) +\mbf{z_i^{(3)}}
    \label{eq:isa_block_apply} 
\end{align}

where $\mathbf{r_i}$ is SAINT's contextual representation output corresponding to data point $\mathbf{x_i}$. This contextual embedding can be used in downstream tasks such as self-supervision or classification.

\subsection{Intersample attention}
\label{subsec:ISA}

\begin{figure}[tb]
  \centering
  \includegraphics[width=\textwidth]{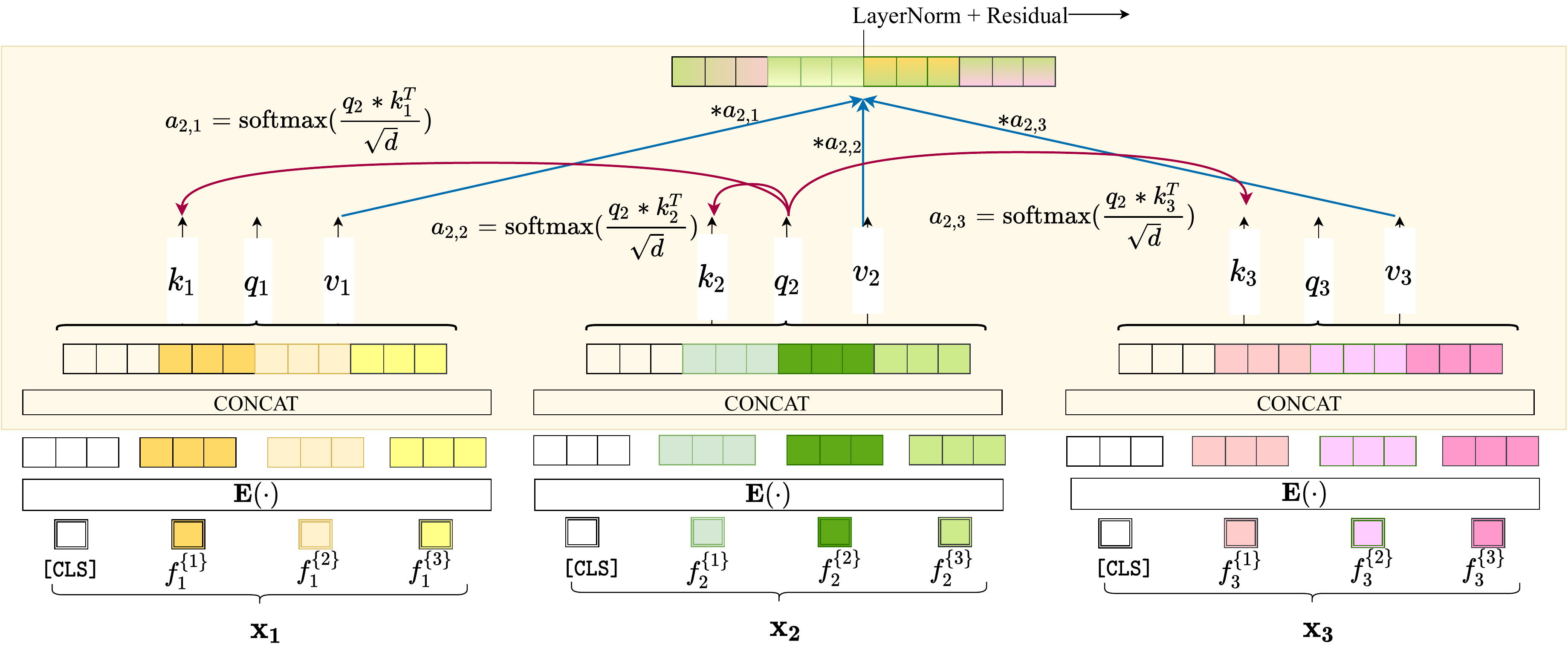}
  \captionsetup{font=small}
  \vspace{-2mm}
  \caption{Intersample attention on a batch of 3 points.  In this plot, $d$ is the size of value vectors $v_i$. See Section~\ref{subsec:ISA} for details. 
  }
  \label{fig:ISA}
\end{figure}
We introduce intersample attention (a type of row attention) where the attention is computed across different data points (rows of a tabular data matrix) in a given batch rather than just the features of a single data point. Specifically, we concatenated the embeddings of each feature for a single data point, then compute attention over samples (rather than features). This enables us to improve the representation of a given point by inspecting other points. 
When a feature is missing or noisy in one row, intersample attention enables SAINT to borrow the corresponding features from other similar data samples in the batch.

An illustration of how intersample attention is performed in a single head is shown in Figure ~\ref{fig:ISA} and the pseudo-code is presented in Algorithm~\ref{alg:MISA}.  Unlike the row attention used in \cite{ho2019axial,child2019generating,rao2021msa,iida2021tabbie}, intersample attention allows all features from different samples to communicate with each other.  In our experiments, we show that this ability boosts performance appreciably.  In the multi-head case, instead of projecting $q,k,v$ to a given dimension $d$, we project them to $d/h$ where $h$ is the number of heads. Then we concatenate all the updated value vectors, $v_i$, to get back a vector of length $d$.

\begin{algorithm}[tb]
\captionsetup{font=small}
   \caption{PyTorch-style pseudo-code for intersample attention. For simplicity, we describe just one head and assume the value vector dimension is same as the input embedding dimension.}
   \label{alg:MISA}
   
    \definecolor{codeblue}{rgb}{0.25,0.5,0.5}
    \lstset{
      basicstyle=\fontsize{8.5 pt}{8.5 pt}\ttfamily\bfseries,
      commentstyle=\fontsize{8.5 pt}{8.5 pt}\color{codeblue},
      keywordstyle=\fontsize{8.5 pt}{8.5 pt}\color{magenta},
    }
    \begin{lstlisting}[language=python]
    # b: batch size, n: number of features, d: embedding dimension 
    # W_q,  W_k, W_v are weight matrices of dimension dxd
    # mm: matrix-matrix multiplication
    def self_attention(x):
        # x is bxnxd
        q, k, v = mm(W_q,x), mm(W_k,x), mm(W_v,x) #q,k,v are bxnxd
        attn = softmax(mm(q,np.transpose(k, (0, 2, 1)))/sqrt(d)) # bxnxn
        out = mm(attn, v) #out is bxnxd
        return out
        
    def intersample_attention(x):
        # x is bxnxd
        b,n,d = x.shape # as mentioned above
        x = reshape(x, (1,b,n*d)) # reshape x to 1xbx(n*d)
        x = self_attention(x) # the output x is 1xbx(n*d)
        out = reshape(x,(b,n,d)) # out is bxnxd
        return out
    \end{lstlisting}
\end{algorithm}

\section{Pre-training \& Finetuning}
\label{sec:pretraining}

Contrastive learning, in which models are pre-trained to be invariant to reordering, cropping, or other label-preserving ``views'' of the data~
\cite{chen2020simple,he2020momentum,pathak2016context,grill2020bootstrap,vincent2008extracting}, is a powerful tool in the vision and language domains that has never (to our knowledge) been applied to tabular data.   
We present a contrastive pipeline for tabular data, a visual description of which is shown in Figure \ref{fig:saint_arch_training}.
Existing self-supervised objectives for tabular data include denoising \citep{vincent2008extracting}, a variation of which was used by VIME~\citep{yoon2020vime}, masking, and replaced token detection as used by TabTransformer \citep{huang2020tabtransformer}. We find that, while these methods are effective, superior results are achieved by contrastive learning.

\textbf{Generating augmentations} Standard contrastive methods in vision craft different ``views'' of images using crops and flips. It is difficult to craft invariance transforms for tabular data.
The authors of VIME~\citep{yoon2020vime} use mixup in the non-embedded space as a data augmentation method, but this is limited to continuous data.
We instead use CutMix~\citep{yun2019cutmix} to augment samples in the input space and we use mixup~\citep{zhang2017mixup} in the embedding space.  
These two augmentations combined yield a challenging and effective self-supervision task.
Assume that only $l$ of $m$ data points are labeled.  We denote the embedding layer by $\mathbf{E}$, the SAINT network by $\mathbf{S}$, and 2 projection heads as $g_1(\cdot)$ and $g_2(\cdot)$. The CutMix augmentation probability is denoted $p_{\text{cutmix}}$ and the mixup parameter is $\alpha$.
Given point $\mathbf{x_i}$, the original embedding is $\mathbf{p_i} = \mathbf{E}(\mathbf{x_i})$, while the augmented representation is generated as follows:
\begin{align}
\mathbf{x_i^\prime} &= \mathbf{x_i}\odot \mathbf{m} + \mathbf{x_a}\odot \mathbf{(1-m)} && \textcolor{gray}{\text{CutMix in raw data space}}\\
\mathbf{p_i^\prime} &= \alpha*\mathbf{E}(\mathbf{x_i^\prime}) +(1-\alpha)*\mathbf{E}(\mathbf{x_b^\prime})&& \textcolor{gray}{\text{mixup in embedding space}}
\end{align}
where $\mathbf{x_a}$, $\mathbf{x_b}$ are random samples from the current batch, $\mathbf{x_b^\prime}$ is the CutMix version of $\mathbf{x_b}$, $\mathbf{m}$ is the binary mask vector sampled from a Bernoulli distribution with probability $p_{\text{cutmix}}$, and $\alpha$ is the mixup parameter. Note that we first obtain a CutMix version of every data point in a batch by randomly selecting a partner to mix with. We then embed the samples and choose new partners before performing mixup.

\textbf{SAINT and projection heads} Now that we have both the clean~$\mathbf{p_i}$ and mixed~$\mathbf{p_i^\prime}$ embeddings, we pass them through SAINT, then through two projection heads, each consisting of an MLP with one hidden layer and a ReLU. The use of a projection head to reduce dimensionality before computing contrastive loss is common in vision~\citep{chen2020simple,he2020momentum,grill2020bootstrap} and indeed also improves results on tabular data. Ablation studies and further discussion are available in Appendix~\ref{appendix_sec:analysis}.


\textbf{Loss functions} We consider two losses for the pre-training phase. (i) The first is a contrastive loss that pushes the latent representations of two views of the same data point ($z_i$ and $z_i^\prime$) close together and encourages different points ($z_i$ and $z_j$, $i\neq j$) to lie far apart.  For this, we borrow the InfoNCE loss from metric-learning works~\citep{sohn2016improved,oord2018representation,chen2020simple, wu2018unsupervised}; (ii) The second loss comes from a denoising task. For denoising, we try to predict the original data sample from a noisy view. Formally, we are given $\mathbf{r_i^\prime}$ and we reconstruct the inputs as $\mathbf{x_i^{\prime \prime}}$ to minimize the difference between the original and the reconstruction.
The combined pre-training loss is:
\begin{align}
\mathcal{L_\text{pre-training}} &= \underbrace{-\sum_{i=1}^{m}{\log{\frac{\exp(z_i \cdot z_i^\prime/\tau)}{\sum_{k=1}^{m}{\exp(z_i \cdot z_k^\prime/\tau)}}}}}_{\text{Contrastive Loss}} + \lambda_{\text{pt}} \underbrace{\sum_{i=1}^{m}\sum_{j=1}^{n} [\mathcal{L}_j(\text{MLP}_j(\mathbf{r_i^\prime}),\mathbf{x_i})]}_{\text{Denoising Loss}}
\end{align}
where $\mathbf{r_i} = \mathbf{S}(\mathbf{p_i}), \mathbf{r_i^\prime} = \mathbf{S}(\mathbf{p_i^\prime}), z_i = g_1(\mathbf{r_i}),z_i^\prime = g_2(\mathbf{r_i^\prime}) $. $\mathcal{L}_j$ is cross-entropy loss or mean squared error depending on the $j^{th}$ feature being categorical or continuous. Each $\text{MLP}_j$ is a single hidden layer perceptron with a ReLU non-linearity. There are $n$ in number, one for each input feature. $\lambda_{\text{pt}}$ is a hyper-parameter and $\tau$ is temperature parameter and both of these are tuned using validation data.

\textbf{Finetuning} Once SAINT is pre-trained on all unlabeled data, we finetune the model on the target prediction task using the $l$ labeled samples.  The pipeline of this step is shown in Figure~\ref{fig:saint_arch_training}(b). For a given point $\mathbf{x_i}$, we learn the contextual embedding $\mathbf{r_i}$. For the final prediction step, we pass the embedding corresponding only to the \verb|[CLS]| token through a simple MLP with a single hidden layer with ReLU activation to get the final output. We evaluate cross-entropy loss on the outputs for classification tasks and mean squared error for regression tasks.

\section{Experimental Evaluation}
\label{sec:expts}
We evaluate SAINT on 16 tabular datasets. In this section, we discuss variants of SAINT and evaluate them in both supervised and semi-supervised scenarios. We also analyze each component of SAINT and perform ablation studies to understand the importance of each component in the model. Using visualization, we interpret the behavior of attention maps.   

\begin{wraptable}{r}{8cm}
\captionsetup{width=8cm,font=small}
    \caption{Configurations of SAINT. The number of stages is denoted by $L$, and the number of heads in each attention layer is represented by $h$. The parameter count is averaged over 14 datasets and is measured for batches of 32 inputs. Time measures the cost of 100 epochs of training plus inference on the best model, averaged over 14 datasets. See Appendix Section~\ref{appendix_sec:complete_training_deets} for hardware specifications.}
    \label{table:SAINT_config}
\resizebox{0.57\textwidth}{!}{
\begin{tabular}{|c|c|c|c|c|c|}
    \hline
    Model & Attention & $L$ & $h$ & Param $\times 1e6$ & Time (s)\\\hline\hline
    SAINT-s & Self & 6 & 8 & 91.6 & 1759\\\hline
    SAINT-i & InterSample & 1 & 8 & 352.7& 123\\\hline
    SAINT & Both & 1 & 8 & 347.3& 144 \\\hline
  \end{tabular}
}
\end{wraptable} 

\textbf{Datasets} We evaluate SAINT on 14 binary classification tasks and 2 multi-class classification tasks. These datasets were chosen because they were previously used to evaluate competing methods \cite{yoon2020vime,arik2019TabNet,huang2020tabtransformer}. They are also diverse; the datasets range in size from 200 to 495,141 samples, and from 8 to 784 features, with both categorical and continuous features. Some datasets are missing data while some are complete and some are well-balanced while others have highly skewed class distributions. Each of these datasets is publicly available from either UCI\footnote{http://archive.ics.uci.edu/ml/datasets.php} or AutoML.\footnote{https://automl.chalearn.org/data} Details of these datasets can be found in Appendix \ref{appendix_sec:datasets}. As the pre-processing step for each dataset, all the continuous features are Z-normalized, and all categorical features are label-encoded before the data is passed on to the embedding layer.

\textbf{Model variants} The SAINT architecture discussed in the previous section has one self-attention transformer encoder block stacked with one intersample attention transformer encoder block in each stage. We also consider variants with just one of these blocks. SAINT-s variant has only self-attention, and SAINT-i has only intersample attention. SAINT-s is exactly the encoder from \cite{vaswani2017attention} but applied to tabular data. See Table \ref{table:SAINT_config} for an architectural comparison of these model variants.


\textbf{Baselines} We compare our model to traditional methods like logistic regression and random forests. We benchmark against the powerful boosting libraries XGBoost, LightGBM, and CatBoost. We also compare against deep learning methods, like multi-layer perceptrons, VIME, TabNet, and TabTransformer. For the methods that use unsupervised pre-training as a preprocessing step, we used Masked Language Modeling~(MLM) for TabNet~\cite{devlin2018bert}, and we use Replaced Token Detection~(RTD) for TabTransformer~\cite{clark2020electra} as mentioned in the respective papers. For multi-layer perceptrons, we use denoising~\cite{vincent2008extracting} as suggested in VIME. 

\textbf{Metrics} Since the majority of the tasks used in our analysis are binary classification, we use AUROC as the primary metric to measure performance. AUROC captures how well the model separates the two classes in the dataset. For the two multi-class datasets, Volkert and MNIST, we use the accuracy on the test set to compare performance.

\textbf{Training} We train all the models (including pre-training runs) using AdamW with $\beta_1 = 0.9$, $\beta_2 = 0.999$, $\text{decay} = 0.01$, and with a learning rate of $0.0001$ with batches of size 256 (except for datasets with a large number of columns like MNIST and Arcene, for which we use smaller batch sizes). We split the data into $65\%$, $15\%$, and $25\%$ for training, validation, and test splits, respectively. We vary the embedding size based on the number of features in the dataset. The exact configurations for each of the datasets are presented in Appendix \ref{appendix_sec:complete_training_deets}. We use CutMix mask parameter $p_{\text{cutmix}} = 0.3$ and mixup parameter $\alpha=0.2$ for all the standard pre-training experiments. We use pre-training loss hyper-parameters $\lambda_{\text{pt}} = 10$ and temperature $\tau = 0.7$ for all settings. 

\subsection{Results}
\label{subsec:main_results}

\begin{table}[t!]
\centering
\captionsetup{font=small}
\caption{Mean AUROC scores (in \%) for SAINT variants and competitors. Results are averaged over 5 trials and 14 binary classification datasets. The mean is over all 14 binary classification datasets. Baseline results are quoted from original papers when possible (denoted with *) and reproduced otherwise.  We highlight best result in bold. Columns denoted by $\dagger$ are multi-class problems, and we report accuracy rather than AUC.}
\resizebox{1\textwidth}{!}{
\begin{tabular}{l | c c c c c c c c c | c } 
 \toprule
 \multicolumn{1}{r|}{Dataset size}   & 45,211  & 7,043     & 452        & 200     & 495,141 & 12,330   & 32,561  & 58,310           & 60,000         &         \\
 \multicolumn{1}{r|}{Feature size} & 16      & 20        & 226        & 783     & 49      & 17       & 14      & 147              & 784            &         \\
 \midrule
Model $\setminus$ Dataset & Bank    & Blastchar & Arrhythmia & Arcene  & Forest  & Shoppers & Income  & Volkert$\dagger$ & MNIST$\dagger$ & Mean    \\
\midrule
Logistic Reg.             & 90.73   & 82.34     & 86.22      & 91.59   & 84.79   & 87.03    & 92.12 & 53.87            & 89.89*         & 89.25   \\
Random Forest             & 89.12 & 80.63   & 86.96    & 79.17 & 98.80 & 89.87  & 88.04 & 66.25            & 93.75          & 89.52 \\
XGBoost~\citep{chen2016xgboost}                   & 92.96 & 81.78   & 81.98    & 81.41 & 95.53 & 92.51  & 92.31 & 68.95            & 94.13*         & 91.06 \\
LightGBM~\citep{ke2017lightgbm}                  & 93.39 & 83.17   & 88.73    & 81.05 & 93.29 & \textbf{93.20}  & \textbf{92.57} & 67.91            & 95.2           & 90.13 \\
CatBoost~\citep{dorogush2018catboost}                  & 90.47 & 84.77   & 87.91    & 82.48 & 85.36 & 93.12  & 90.80 & 66.37            & 96.6           & 90.73 \\
MLP                       & 91.47   & 59.63     & 58.82      & 90.26   & 96.81   & 84.71    & 92.08   & 63.02            & 93.87*         & 84.59    \\
VIME~\citep{yoon2020vime}                      & 76.64   & 50.08     & 65.3       & 61.03   & 75.06   & 74.37    & 88.98   & 64.28            & 95.77*         & 76.07   \\
TabNet~\citep{arik2019TabNet}                    & 91.76 & 79.61   & 52.12    & 54.10 & 96.37 & 91.38  & 90.72 & 56.83            & 96.79          & 83.88   \\
TabTransf.~\citep{huang2020tabtransformer}            & 91.34   & 81.67     & 70.03      & 86.8    & 84.96   & 92.70*   & 90.60*  & 57.98            & 88.74          & 90.86   \\
\midrule
SAINT-s                   & \textbf{93.61}   & \textbf{84.91}     & 93.46      & 86.88   & 99.67   & 92.92    & 91.79   & 62.91            & 90.52          & 92.59   \\
SAINT-i                   & 92.83   & 84.46     & \textbf{95.8}       & \textbf{92.75}   & 99.45   & 92.29    & 91.55   & \textbf{71.27}            & \textbf{98.06 }         & 93.09   \\
SAINT                     & 93.3    & 84.67     & 94.18      & 91.04   & \textbf{99.7 }   & 93.06    & 91.67   & 70.12            & 97.67          & \textbf{93.13}   \\
\bottomrule
\end{tabular}
}

\label{table:supervised_results}
\end{table}

\textbf{Supervised setting} In Table~\ref{table:supervised_results}, we report results on 7 representative binary classification and 2 multi-class classification datasets, as well as the average performance across all 14 binary classification datasets. Note that each number reported in the Table~\ref{table:supervised_results} is the mean from 5 trials with different seeds. In 13 out of 16 datasets, one of the SAINT variants outperforms all baseline models. In the remaining 3 datasets, our model's performance is very close to the best model. On average, SAINT variants each perform better than all baseline models by a significant margin, and SAINT performs even better than its two variants. For complete results from every dataset as well as details including standard error, see Appendix \ref{appendix_sec:exhaustive_results}.

\textbf{Semi-supervised setting}  We perform 3 sets of experiments with 50, 200, and 500 labeled data points (in each case the rest are unlabeled). See Table~\ref{table:semisupervised} for numerical results. In all cases, the pre-trained SAINT model (with both self and intersample attention) performs the best. Interestingly, we note that when all the training data samples are labeled, pre-training does not contribute appreciably, hence the results with and without pre-training are fairly close.

\textbf{Effect of embedding continuous features} To understand the effect of learning embeddings for continuous data, we perform a simple experiment with TabTransformer. We modify TabTransformer by embedding continuous features 
into $d$ dimensions using a single layer ReLU MLP, just as they use on categorical features, and we pass the embedded features through the transformer block. We keep the entire architecture and all training hyper-parameters the same for both TabTransformer and its modified version. The average AUROC of the original TabTransformer is 89.38. Just by embedding the continuous features, the performance jumps to 91.72. This experiment shows that embedding the continuous data is important and can boost the performance of the model significantly. 

\begin{wraptable}{r}{7cm}
\captionsetup{width=7cm,font=small}
    \captionof{table}{Average AUROC scores (in \%) across 14 datasets under semi-supervised scenarios. Columns vary by number of labeled training samples. The last column is a repetition of results from Table \ref{table:supervised_results}.}
    \label{table:semisupervised}
\resizebox{0.5\textwidth}{!}{
\begin{tabular}{l | c c c c  } 
 \toprule
 Model $\setminus$ \# Labeled  & 50 & 200 & 500 & All \\
 \midrule
Logistic Reg. & 78.69  & 78.93 & 82.13 & 89.25    \\ 
Random Forest& 80.99  & 82.56 & 86.76  & 89.52   \\ 
XGBoost~\cite{chen2016xgboost} & 78.13  & 83.89 & 86.20  &91.06    \\ 
LightGBM~\cite{ke2017lightgbm} & 78.69 & 85.60 & 87.94  & 90.13   \\ 
CatBoost~\cite{dorogush2018catboost} & 84.08  & 88.08 & 88.84  & 90.73  \\ 
MLP w. DAE~\cite{vincent2008extracting} & 71.59  & 78.84 & 81.98  & 84.59    \\ 
VIME~\cite{yoon2020vime} & 67.22  & 74.43 & 76.24  & 76.07    \\ 
TabNet w. MLM~\cite{arik2019TabNet} & 67.31  & 71.72 & 76.01  & 83.88   \\ 
TabTransf. w. RTD~\cite{huang2020tabtransformer} & 82.41 & 86.16 & 87.36  & 90.86    \\ 
\midrule
SAINT-s  & 85.14  & 87.32 & 88.89  & 92.59   \\ 
SAINT-i & 83.93  & 84.65 & 88.12  & 93.09   \\ 
SAINT & 85.78  & 87.66 & 89.12  & \textbf{93.13}   \\ 
\midrule
SAINT-s  + pre-training & 85.92  & 87.94 & 89.19  & 92.64    \\ 
SAINT-i + pre-training & 84.88  & 88.41 & 88.77 & 93.02   \\ 
SAINT+ pre-training & \textbf{86.91} & \textbf{88.69} & \textbf{89.22}  & 92.99      \\ [1ex] 
\bottomrule
\end{tabular}
}
\end{wraptable} 

\textbf{When to use intersample attention?} From our experiments, we observe that SAINT-i consistently outperforms other variants whenever the number of features is large. In particular, whenever there are few training data points coupled with many features (which is common in biological datasets), SAINT-i outperforms SAINT-s significantly (see the ``Arcene'' and ``Arrhythmia'' results). Another advantage of using SAINT-i is that execution is fast compared to SAINT-s,  despite the fact that the number of parameters of SAINT-i is much higher than that of SAINT-s (see Table~\ref{table:SAINT_config}). 

\textbf{How robust is SAINT to data corruptions?} We evaluate the robustness of SAINT variants by corrupting the training data.  To simulate corruption, we apply CutMix, replacing 10\% to 90\% of the features with values of other randomly selected samples. The drop in the mean AUROC is quite minimal until 70\% data corruption when the performance drops significantly. SAINT and SAINT-i models are comparatively more robust than SAINT-s. This shows that using row attention improves the model's robustness to noisy training data as we anticipated. However, we find the opposite trend when many features are missing in the training data. SAINT-s and SAINT are quite robust, and the drop in AUROC is not drastic even when 90\% of the data is missing. This observation shows that SAINT is reliable for training on corrupted training data. The AUROC trend line plots for both the scenarios are shared in Appendix \ref{appendix_sec:analysis}.

\textbf{Effect of batch size on intersample attention performance} As discussed in Section~\ref{subsec:ISA}, attention is computed between batches of data points. We examine the impact of batch size using batches of size ranging from 32 to 256. We find that the variation in SAINT-i's performance is low and is comparable to that of SAINT-s, which has no intersample attention component. We present the plots in Appendix \ref{appendix_sec:analysis}.

\subsection{Interpreting attention}
\label{subsec:attention_analysis}
One advantage of using transformer-based models is that attention comes with some interpretability, in contrast, MLPs are hard to interpret. In particular, when we use only one transformer stage, the attention maps reveal which features and which data points are being used by the model to make decisions. We use MNIST data to examine how self-attention and intersample attention behave in our models. While MNIST is not a typical tabular dataset, it has the advantage that its features can be easily visualized as an image.

\begin{figure}
     \centering
     \begin{subfigure}[b]{0.3\textwidth}
         \centering
         ~\vspace{4mm}
         \includegraphics[width=\textwidth,valign=t]{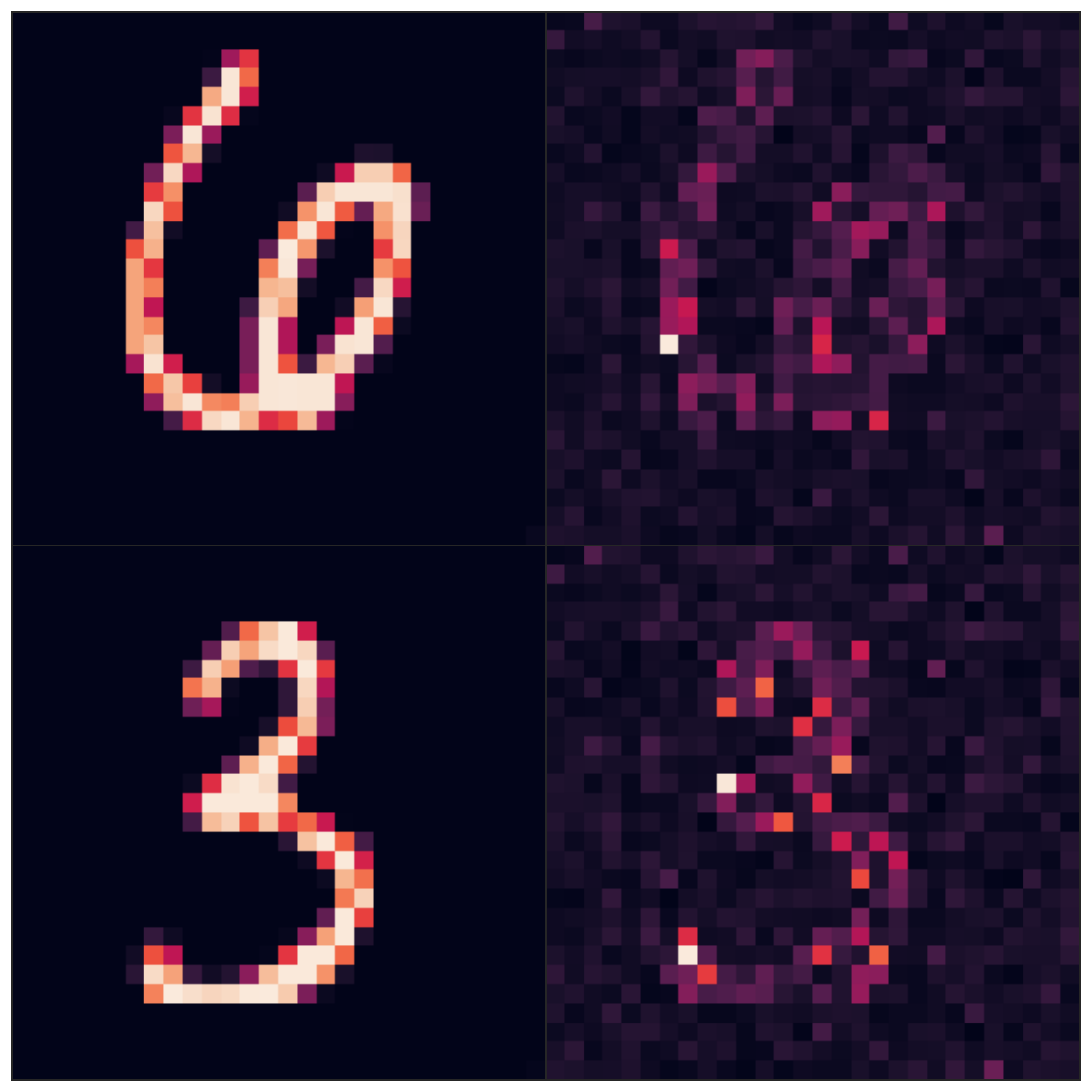}
         \caption{Self-attn. in SAINT }
         \label{fig:SA_in_SAINT}
     \end{subfigure}
     \hfill
     \begin{subfigure}[b]{0.32\textwidth}
         \centering
         \includegraphics[width=\textwidth,valign=t]{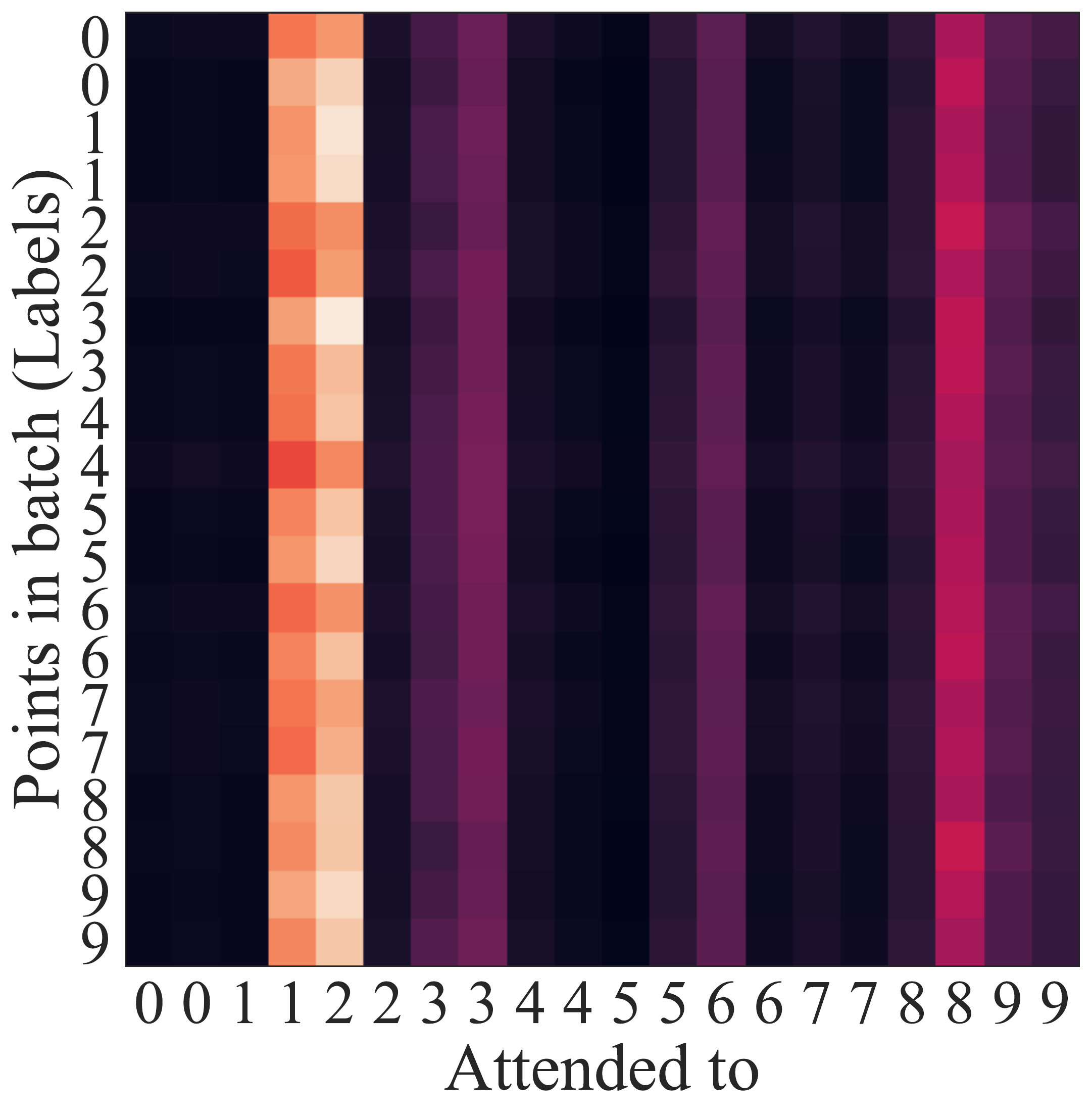}
         \caption{Intersample attn. in SAINT}
         \label{fig:ISA_in_SAINT}
     \end{subfigure}
     \hfill
     \begin{subfigure}[b]{0.32\textwidth}
         \centering
         \includegraphics[width=\textwidth,valign=t]{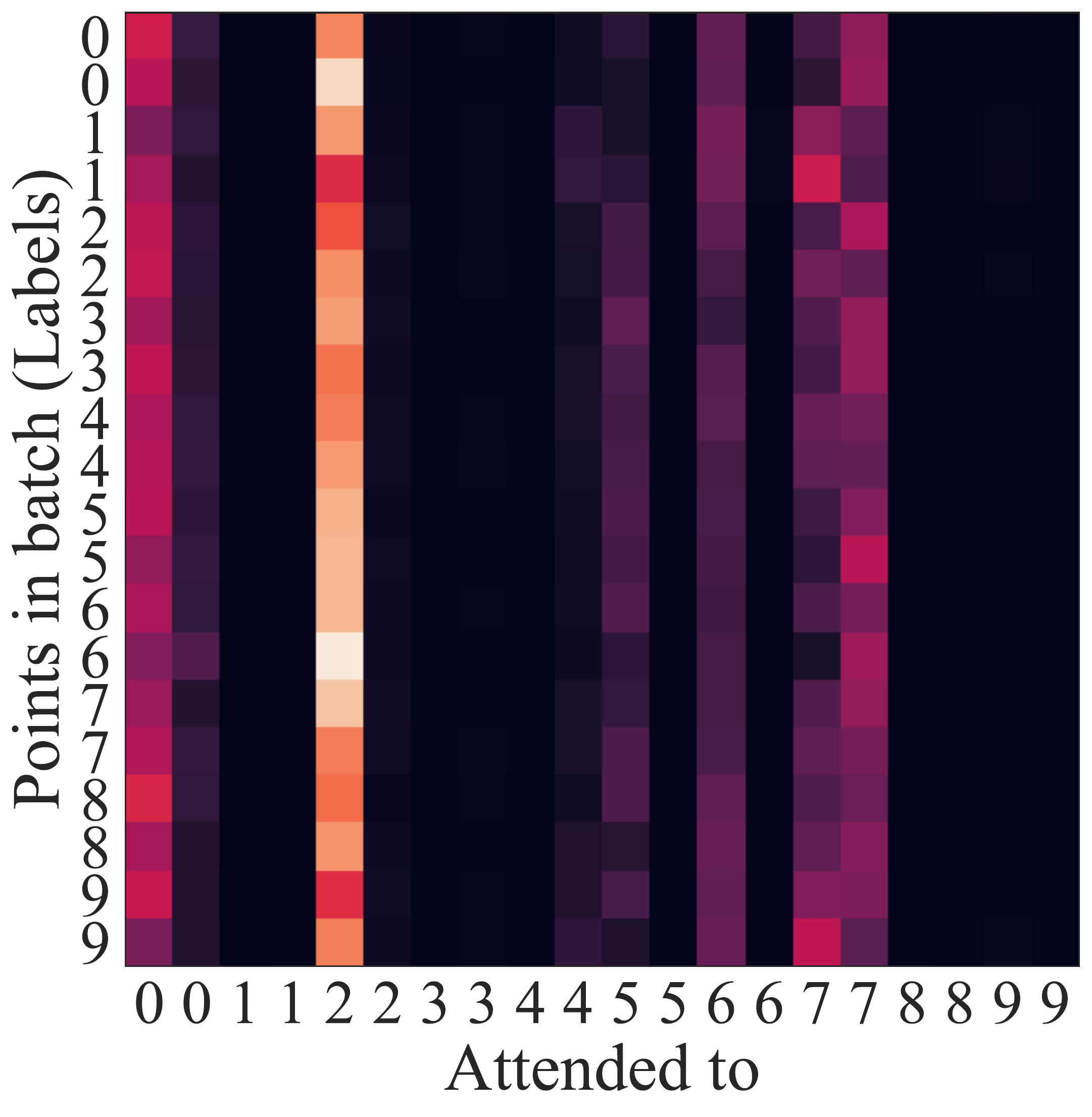}
         \caption{Intersample attn. in SAINT-i}
         \label{fig:ISA_in_SAINTi}
     \end{subfigure}
     \captionsetup{font=small}
        \caption{Visual representations of various attention mechanisms. }
        \label{fig:attention_plots}
\end{figure}

Figure~\ref{fig:SA_in_SAINT} depicts the attention on each of the pixels/features in a self-attention layer of SAINT. Without any explicit supervision, the model learns to focus on the foreground pixels, and we clearly see from the attention map which features are most important to the model. The self-attention plots of SAINT-s are similar (Appendix \ref{appendix_sec:additional_interpret}).

Figures~\ref{fig:ISA_in_SAINT} and \ref{fig:ISA_in_SAINTi} depict a similar visualization on a batch of 20 points, 2 from each class in MNIST. Figure~\ref{fig:ISA_in_SAINT} shows intersample attention in SAINT. This plot shows which samples attend to which other samples in the batch. Surprisingly, very few points in a batch receive attention. We hypothesize that the model focuses on a few points that are critical because they are particularly difficult to classify without making direct comparisons to exemplars in the batch.
In Figure~\ref{fig:ISA_in_SAINTi}, we show the intersample attention plot from a SAINT-i model. The same sparse attention behaviour persists here too, but the points being attended to are different in this model. Interestingly, we find this behavior to be significantly different on the Volkert data, where a wide range of data becomes the focus of attention depending on the input. The intersample attention layer gets dense with the hardness~(to classify) of the datasets. See Appendix ~\ref{appendix_sec:additional_interpret} for additional MNIST and Volkert attention maps.

\begin{figure}
    \centering
     \begin{subfigure}[b]{0.45\textwidth}
         \centering
         \includegraphics[width=\textwidth]{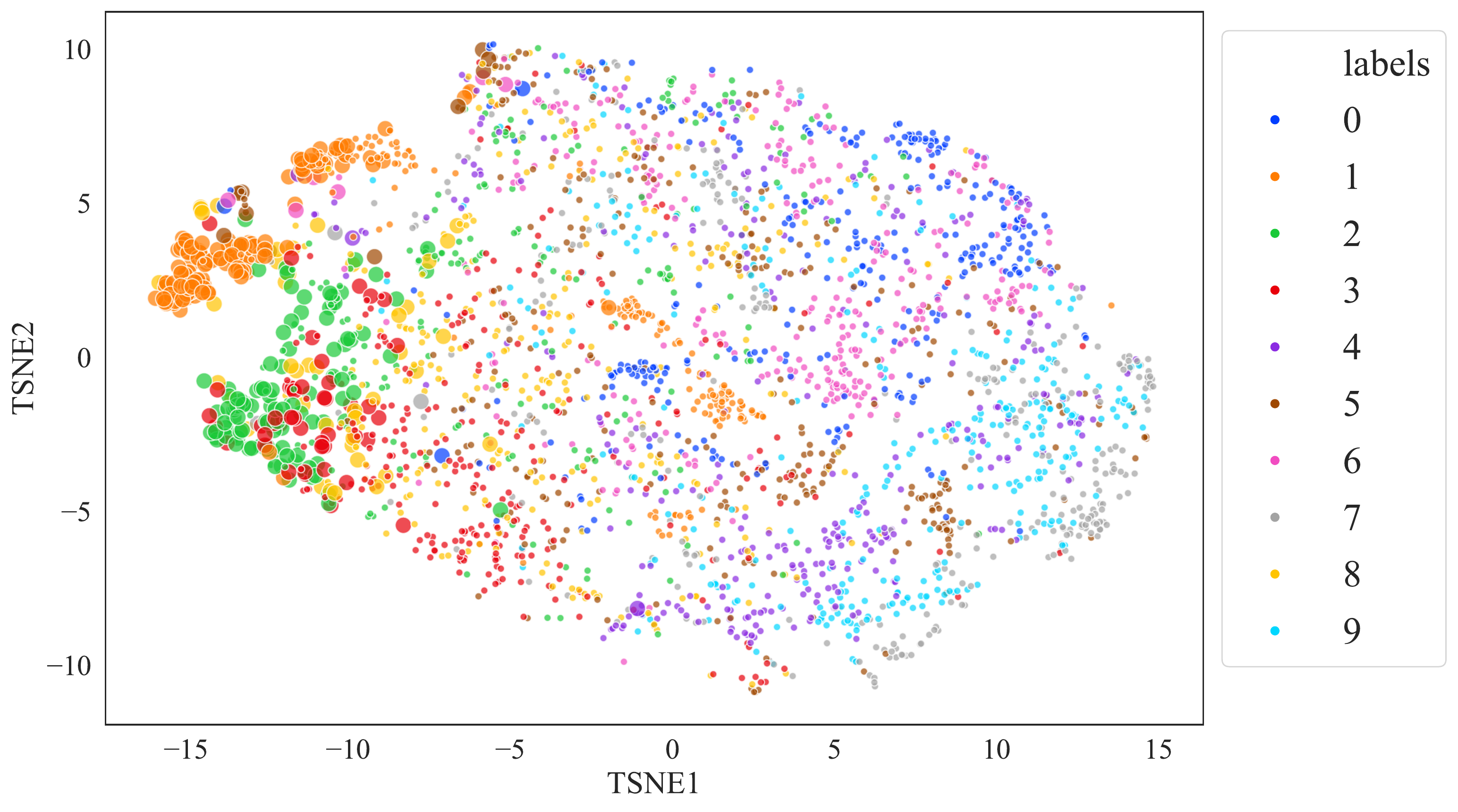}
         \label{fig:saint_tsne}
     \end{subfigure}
     \begin{subfigure}[b]{0.45\textwidth}
         \centering
         \includegraphics[width=\textwidth]{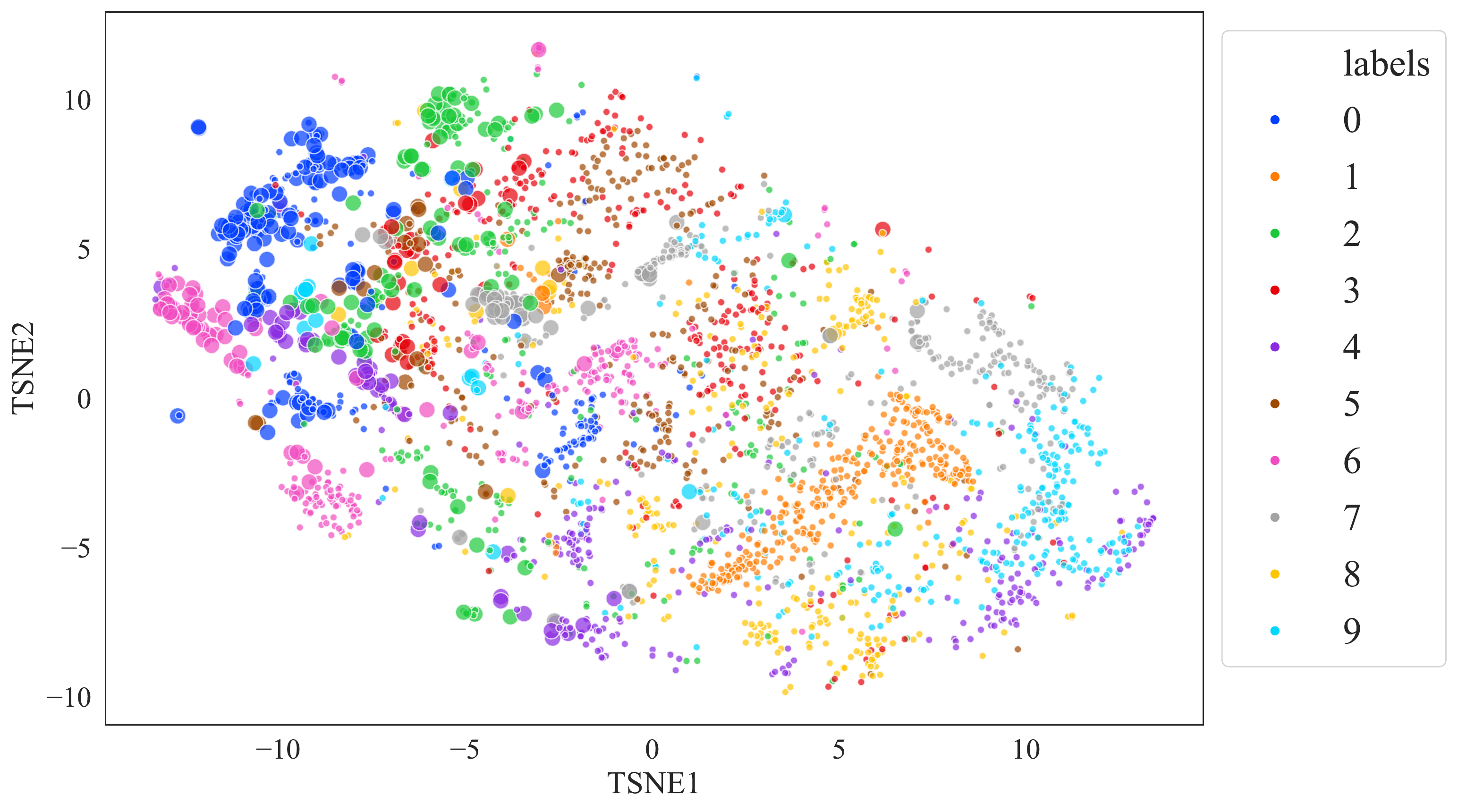}
         \label{fig:saint_i_tsne}
     \end{subfigure}
     \captionsetup{font=small}
     \vspace{-5mm}
    \caption{A t-SNE visualization of \textit{value} vectors in intersample attention layers of SAINT~(left) and SAINT-i~(right). We plot 3500 points in each figure, with classes uniformly represented. In the left plot, we  observe that the most attended classes are 1, 2, 3, and 8. But in the right plot, the most attended classes are 0, 2, 6, and 7.}
    \label{fig:attention_tsne}
\end{figure}

Figure~\ref{fig:attention_tsne} shows the behavior of attention at the dataset (rather than batch) level. We visualize a t-SNE \citep{van2008visualizing} embedding for \textit{value} vectors generated in intersample attention layers, and we highlight the points that are most attended to in each batch. In Figure~\ref{fig:attention_tsne}~(left), the \textit{value} vectors and attention are computed on the output representations of a self-attention layer. 
In contrast, the \textit{value} vectors and attention in Figure~\ref{fig:attention_tsne}~(right) are computed on the embedding layer output, since the SAINT-i model does not use self-attention.  In these two plots, the classes to which the model attends vary dramatically. Thus, the exact classes to which an attention head pays attention change with the architecture, but the trend of using a few classes as a `pivot' seems to be prevalent in intersample attention heads. 
Additional analyses are presented in Appendix \ref{appendix_sec:analysis}.


\section{Discussion, Limitations, and Impact}
\label{sec:discussion}
We introduce intersample attention, contrastive pre-training, and an improved embedding strategy for tabular data. Even though tabular data is an extremely common data format used by institutions in various domains, deep learning methods are still lagging behind tree-based boosting methods in production. With SAINT, we show that neural models can often improve upon the performance of boosting methods across numerous datasets with varying characteristics.

SAINT offers improvements in a widely used domain, which is quite impactful. While our method performs well on the diverse tabular datasets studied here, real-world applications contain a broad range of datasets which may be highly noisy or imbalanced.  Moreover, we have tuned SAINT for the settings in which we test it.  Thus, we caution practitioners against assuming that what works on the benchmarks in this paper will work in their own setting.

{
\small
\bibliographystyle{plainnat}
\bibliography{references}
}

\newpage

\clearpage
\appendix
\textbf{\Large{Appendix for SAINT: Improved Neural Networks for Tabular Data via Row Attention and Contrastive Pre-Training}}

\section{Additional illustrations}
\label{appendix_sec:extra_illustrations}

\begin{figure}[h]
  \centering
  \includegraphics[width=\textwidth]{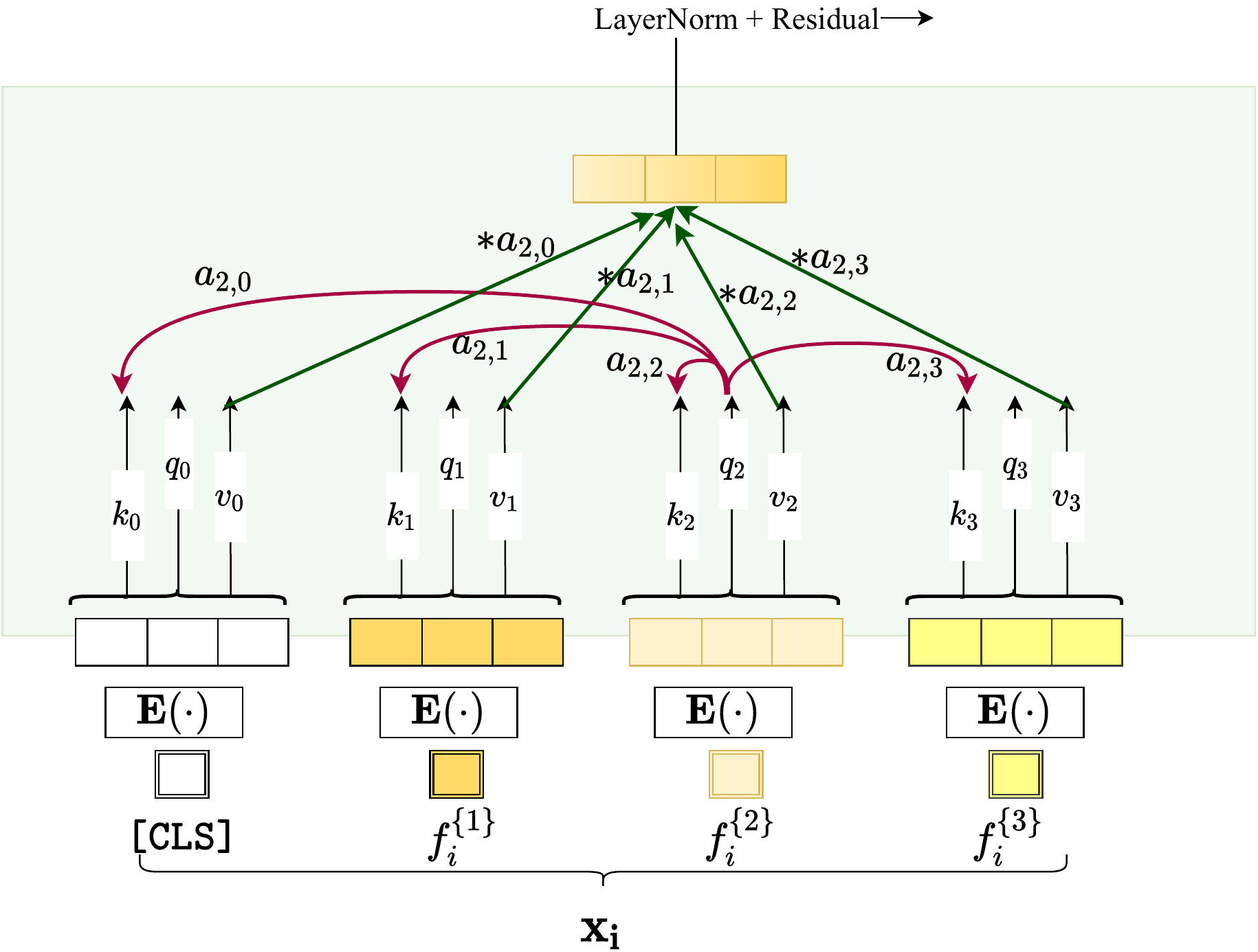}
  \caption{An illustration of self-attention in a point $\mathbf{x_i}$. Inspired by Vaswani et al~\cite{vaswani2017attention}.}
  \label{fig:self_attention_illustration}
\end{figure}

\section{Datasets}
\label{appendix_sec:datasets}

\textbf{Data sources} For each dataset, details and download links are listed in Tables \ref{tab:dataset_details} and \ref{tab:datasetlinks}. The 1995 Income Classification dataset is from a 2019 Kaggle competition and was made public without a license. The Arcene dataset, furnished by UCI \cite{Dua:2019}, comprises anonymized patient records, where the goal is to classify entries as containing cancer patterns or normal patterns \cite{guyon2004result}. The Arrhythmia dataset is made available by Stonybrook University \cite{liu2008isolation, ting2009mass, keller2012hics}. The Bank Marketing dataset is also compiled and organized by UCI and released for research use \cite{moro2014data}. The BlastChar dataset is fictitious, it is also part of a Kaggle competition and was originally generated by IBM \cite{ibm_blastchar}. The Credit Card dataset is provided through another Kaggle competition under the CC0 license for public use. The Forest data is available through UCI and was originally donated to their archive from Colorado State University in Fort Collins. It is copyrighted by Jock A. Blackard and Colorado State University but available for unlimited use. The HTRU2 dataset is also available through the UCI archive and is available for research use \cite{lyon2016fifty}. The KDD 99 data consists of digital connection data where the task is to classify connections as good or bad, thereby detecting intrusions \cite{stolfo2000cost}. Online Shoppers data is available through UCI and is designed to capture the difference between the behavior of online shoppers who make a purchase and those who do not \cite{sakar2019real}. The Philippine dataset is available through AutoML for research use and does not have a license. The QSAR data is also available through the UCI archive \cite{mansouri2013quantitative}. The Shrutime consists of anonymized bank records that can be used to determine whether a customer closed their account at that bank. It is made available without a license for a Kaggle competition. The Spambase data was originally compiled by Hewlett-Packard and donated to the UCI archive. The Volkert data is available through AutoML. The MNIST data is available at the link provided \cite{lecun1998gradient}.

\begin{table}[h!]
\centering
\captionsetup{font=small}
\caption{We present statistics on 16 datasets we have used in this paper, 14 of which involve binary classification and 2 of which involve multiclass classification (10 classes).}
\label{tab:dataset_details}
\resizebox{1\textwidth}{!}{
\begin{tabular}{l | c c c c c c c c } 
\toprule
Dataset          & Task       & \# Features & \# Categorical & \# Continuous & Dataset Size & \# Positives & \# of Neg. & \% of Positives \\ 
\midrule
Income     & Binary     & 14        & 8           & 6          & 32,561       & 7,841           & 24,720    & 24.08   \\
Arcene           & Binary     & 783       & 0           & 783        & 200          & 88              & 112       & 44.00   \\
Arrhythmia       & Binary     & 226       & 0           & 226        & 452          & 66              & 386       & 14.60   \\
Bank  & Binary     & 16        & 9           & 7          & 45,211       & 5,289           & 39,922    & 11.70   \\
BlastChar        & Binary     & 20        & 17          & 3          & 7,043        & 1,869           & 5,174     & 26.54   \\
Credit       & Binary     & 29        & 0           & 29         & 284,807      & 492             & 284,315   & 0.17    \\
Forest           & Binary     & 49        & 0           & 49         & 495,141      & 283,301         & 211,840   & 57.22   \\
HTRU2            & Binary     & 8         & 0           & 8          & 17,898       & 1,639           & 16,259    & 9.16    \\
KDD99            & Binary     & 39        & 3           & 36         & 494,021      & 97,278          & 396,743   & 19.69   \\
Shoppers  & Binary     & 17        & 2           & 15         & 12,330       & 1,908           & 10,422    & 15.47   \\
Philippine       & Binary     & 308       & 0           & 308        & 5,832        & 2,916           & 2,916     & 50.00   \\
QSAR Bio        & Binary     & 41        & 0           & 41         & 1,055        & 356             & 699       & 33.74   \\
Shrutime         & Binary     & 11        & 3           & 8          & 10,000       & 2,037           & 7,963     & 20.37   \\
Spambase         & Binary     & 57        & 0           & 57         & 4,601        & 1,813           & 2,788     & 39.40   \\
Volkert          & Multiclass (10) & 147       & 0           & 147        & 58,310       & -               & -         & -   \\
MNIST            & Multiclass (10) & 784       & 784         & 0          & 60,000       & -               & -         & -   \\
\bottomrule
\end{tabular}
}
\end{table}

\begin{table}[h!]
\centering
\captionsetup{font=small}
\caption{Dataset links}
\label{tab:datasetlinks}
\resizebox{1\textwidth}{!}{
\begin{tabular}{l | l } 
\toprule
Dataset          & Download Link \\
\midrule
Income     & \url{https://www.kaggle.com/lodetomasi1995/income-classification}                          \\
Arcene           & \url{https://archive.ics.uci.edu/ml/machine-learning-databases/arcene/}                    \\
Arrhythmia       & \url{http://odds.cs.stonybrook.edu/arrhythmia-dataset/}                                    \\
Bank  & \url{https://archive.ics.uci.edu/ml/datasets/bank+marketing}                               \\
BlastChar        & \url{https://www.kaggle.com/blastchar/telco-customer-churn}                                \\
Credit       & \url{https://www.kaggle.com/jacklizhi/creditcard}                                          \\
Forest           & \url{https://kdd.ics.uci.edu/databases/covertype}                                          \\
HTRU2            & \url{https://archive.ics.uci.edu/ml/datasets/HTRU2}                                        \\
KDD 99            & \url{http://kdd.ics.uci.edu/databases/kddcup99}                                            \\
Shoppers & \url{https://archive.ics.uci.edu/ml/datasets/Online+Shoppers+Purchasing+Intention+Dataset} \\
Philippine       & \url{http://automl.chalearn.org/data}                                                      \\
QSAR Bio        & \url{https://archive.ics.uci.edu/ml/datasets/QSAR+biodegradation}                          \\
Shrutime         & \url{https://www.kaggle.com/shrutimechlearn/churn-modelling}                               \\
Spambase         & \url{https://archive.ics.uci.edu/ml/datasets/Spambase}                                     \\
Volkert          & \url{http://automl.chalearn.org/data}                                                      \\
MNIST            & \url{http://yann.lecun.com/exdb/mnist/}     \\
\bottomrule
\end{tabular}
}
\end{table}

\paragraph{Data preprocessing}
In each dataset, the categorical features are label encoded, and continuous features are z-normalized before passing them into the embedding layer. Each feature (or column) has a different missing value token to account for missing data. Additionally, individual datasets contain the following assumptions. In the Arcene, Arrhythmia, and KDD99 datasets, many features have identical values across samples (i.e. zero standard deviation), so we have removed these features. In the Forest dataset, following \cite{arik2019TabNet}, we have considered only the top 2 classes as a binary classification problem.

For MNIST, we unravel each image into a vector of 784 features and consider each image as a single row. Since each feature is of same type in this dataset, we encode all the features into the same embedding space. To distinguish the features, we also use positional encodings in the encoding layer.

\section{Complete training details}
\label{appendix_sec:complete_training_deets}

In each of our experiments, we use a single Nvidia GeForce RTX 2080Ti GPU. Individual training runs take between 5 minutes and 6 hours. In total, the experiments in this paper account for around 4 GPU days (including semi-supervised experiments and ablation studies).

For most of the datasets, we use embedding size $d=32$. For MNIST, we use $d=12$, for the Arrhythmia, Philippine, and Credit datasets we used $d=8$, for Arcene we use $d=4$. The variance in the embedding size is only due to the memory constraints of a single GPU. We used $L=6$ layers in the SAINT-s variant for most of the datasets except for Arrhythmia, Philippine and Arcene, where we use $L=4$ due to memory constraints. We use dropout of 0.1 in all attention layers. In feed-forward layers, use dropout of 0.1 in the SAINT-s variant, and we use 0.8 in SAINT-i and SAINT models. We use attention heads $h=8$ in all datasets except Arrhythmia, Philippine, Credit, Arcene, and MNIST where we use $h=4$ since we are using a lower embedding size. Inside the self-attention layer, the $q$, $k$, and $v$ vectors are of dimension 16, and in the intersample attention layer, they are of size 64.

Other minor details are shared in the code.

\paragraph{Positional Encoding} Transformers for vision and language typically employ positional encodings along with the patch/word embeddings to retain spatial information. These encodings are necessary when all features in a data point are of same type, hence these models use the same function to embed all inputs. This is not the case with most of the datasets used in this paper; each feature may be of a different type and thus possesses a unique embedding function. However, when we train the model on MNIST (treated as tabular data), positional encodings are used since all pixels are of the same type and share a single embedding function.

\section{Additional results}
\label{appendix_sec:exhaustive_results}
\paragraph{Standard errors of datasets shown in main}
In Table~\ref{table:std_errors_1}, we include standard errors on AUROC scores across the various datasets shared in the main document. We see that Arrhythmia and Arcene have high standard error across all models which can be attributed to the size of the datasets (400 and 200 datapoints respectively). Boosting methods are more consistent than previous deep learning approaches, but SAINT's variants exhibit the same consistency as boosting methods.

\paragraph{Remaining datasets} In Table~\ref{table:auroc_results_2}, we share the average AUROC scores over 5 runs for the remaining 7 binary classification datasets which are not shown in the main paper. In Table~\ref{table:std_errors_2}, we show the standard errors over these 7 datasets.

\begin{table}[h!]
\centering
\captionsetup{font=small}
\caption{Std. errors on AUROC scores (in \%) for SAINT variants and competitors. Computed over 5 runs. Columns denoted by $\dagger$ are multi-class problems, and we report standard errors (over 2 runs) on accuracy rather than AUC.}
\resizebox{0.9\textwidth}{!}{
\begin{tabular}{l|ccccccccc}
\toprule
Model \textbackslash~Dataset & Bank   & Blastchar & Arrhythmia & Arcene & Forest & Shoppers & Income & Volkert$\dagger$ & MNIST$\dagger$  \\
\midrule
Logistic Regression          & 0.25 & 0.20    & 2.92     & 2.43 & 0.11 & 0.41   & 6.34 & 1.33  & 3.19 \\
RandomForest                 & 0.27 & 0.70    & 1.51     & 3.29 & 0.01 & 0.60   & 0.30 & 1.27  & 4.59 \\
XGBoost                      & 0.15 & 0.34    & 3.03     & 1.91 & 0.01 & 0.50   & 0.15 & 0.51  & 1.98 \\
LightGBM                     & 0.21 & 0.34    & 1.98     & 1.11 & 0.01 & 0.48   & 0.13 & 0.64  & 3.78 \\
CatBoost                     & 0.17 & 0.19    & 2.60     & 1.62 & 0.01 & 0.41   & 0.15 & 1.17  & 1.66 \\
MLP                          & 0.21 & 0.32    & 2.76     & 3.46 & 0.68 & 0.60   & 2.74 & 1.56  & 3.74 \\
VIME                         & 2.03 & 0.26    & 2.14     & 3.45 & 6.91 & 2.74   & 5.10 & 6.67  & 8.15 \\
TabNet                       & 0.33 & 0.30    & 6.38     & 2.72 & 0.01 & 0.68   & 0.17 & 1.47  & 2.22 \\
Tabtransformer               & 0.34 & 0.30    & 6.45     & 2.75 & 0.01 & 0.69   & 0.17 & 1.48  & 2.24 \\
\midrule
SAINT-s                      & 0.15 & 0.39    & 1.49     & 2.07 & 0.00 & 0.33   & 0.24 & 0.49  & 1.71 \\
SAINT-i                      & 0.09 & 0.22    & 3.37     & 1.78 & 0.02 & 0.42   & 0.24 & 0.67  & 1.49 \\
SAINT                        & 0.09 & 0.28    & 1.94     & 1.41 & 0.01 & 0.30   & 0.27 & 0.58  & 1.13 \\
\bottomrule
\end{tabular}
}
\label{table:std_errors_1}
\end{table}

\begin{table}[t!]
\centering
\captionsetup{font=small}
\caption{Average AUROC scores (in \%) for SAINT variants and competitors on 7 the remaining binary classification datasets. Computed over 5 runs.}
\resizebox{0.8\textwidth}{!}{
\begin{tabular}{l|ccccccc}
\toprule
Model \textbackslash Dataset & Credit  & HTRU2   & QSAR Bio & Shrutime & Spambase & Philippine & KDD99 \\
\midrule
Logistic Regression          & 96.85 & 98.23 & 84.06 & 83.37  & 92.77  & 79.48    & 99.98  \\
Random Forest                & 92.66 & 96.41 & 91.49 & 80.87  & 98.02  & 81.29    & \textbf{100.00} \\
XGBoost                      & \textbf{98.20} & 97.81 & 92.70 & 83.59  & 98.91  & \textbf{85.15}    & \textbf{100.00} \\
LightGBM                     & 76.07 & 98.10 & 92.97 & 85.36  & \textbf{99.01}  & 84.97    & \textbf{100.00} \\
CatBoost                     & 96.83 & 97.85 & 93.05 & 85.44  & 98.47  & 83.63    & \textbf{100.00} \\
MLP                          & 97.76 & 98.35 & 79.66 & 73.70  & 66.74  & 79.70    & 99.99  \\
VIME                         & 82.63 & 97.02 & 81.04 & 70.24  & 69.24  & 73.51    & 99.89  \\
TabNet                       & 95.24 & 97.58 & 67.55 & 75.24  & 97.93  & 74.21    & \textbf{100.00} \\
Tab Transformer              & 97.31 & 96.56 & 91.80 & 85.60  & 98.50  & 83.40    & \textbf{100.00} \\
\midrule
SAINT-s                      & 98.08 & 98.16 & 92.89 & 86.40  & 98.21  & 79.30    & \textbf{100.00} \\
SAINT-i                      & 98.12 & \textbf{98.36} & \textbf{93.48} & 85.68  & 98.40  & 80.08    & \textbf{100.00 }\\
SAINT                        & 97.92 & 98.08 & 93.21 & \textbf{86.47}  & 98.54  & 81.96    & \textbf{100.00} \\
\bottomrule
\end{tabular}
}
\label{table:auroc_results_2}
\end{table}

\begin{table}[h!]
\centering
\captionsetup{font=small}
\caption{Std. errors on AUROC (in \%) scores for SAINT variants and competitors on the 7 remaining binary classification datasets. Computed over 5 runs.}
\resizebox{0.8\textwidth}{!}{
\begin{tabular}{l|ccccccc}
\toprule
Model \textbackslash Dataset & Credit & HTRU2  & QSAR Bio & Shrutime & Spambase & Philippine & KDD99  \\
\midrule
Logistic Regression          & 0.61 & 0.26 & 0.70  & 0.53   & 0.12   & 0.09     & 0.00 \\
RandomForest                 & 0.87 & 0.25 & 0.80  & 0.38   & 0.27   & 0.09     & 0.00 \\
XGBoost                      & 0.38 & 0.10 & 0.45  & 0.39   & 0.08   & 0.09     & 0.00 \\
LightGBM                     & 0.72 & 0.13 & 0.67  & 0.58   & 0.05   & 0.14     & 0.00 \\
CatBoost                     & 0.31 & 0.23 & 0.79  & 0.41   & 0.11   & 0.31     & 0.00 \\
MLP                          & 0.71 & 0.31 & 1.00  & 1.65   & 0.15   & 0.84     & 0.00 \\
VIME                         & 2.18 & 2.52 & 0.71  & 1.15   & 3.03   & 4.67     & 0.00 \\
TabNet                       & 0.42 & 0.29 & 2.67  & 5.12   & 0.15   & 1.21     & 0.00 \\
Tabtransformer               & 0.43 & 0.29 & 2.70  & 5.18   & 0.15   & 1.23     & 0.00 \\
\midrule
SAINT-s                      & 0.32 & 0.10 & 0.81  & 0.68   & 0.21   & 0.16     & 0.00 \\
SAINT-i                      & 0.28 & 0.13 & 1.04  & 0.58   & 0.14   & 0.20     & 0.00 \\
SAINT                        & 0.21 & 0.12 & 0.91  & 0.52   & 0.29   & 0.40     & 0.00 \\
\bottomrule
\end{tabular}
}
\label{table:std_errors_2}
\end{table}

\section{Additional analyses}
\label{appendix_sec:analysis}

\begin{figure}[t]
     \centering
     \begin{subfigure}[b]{0.48\textwidth}
         \centering
         \includegraphics[width=\textwidth]{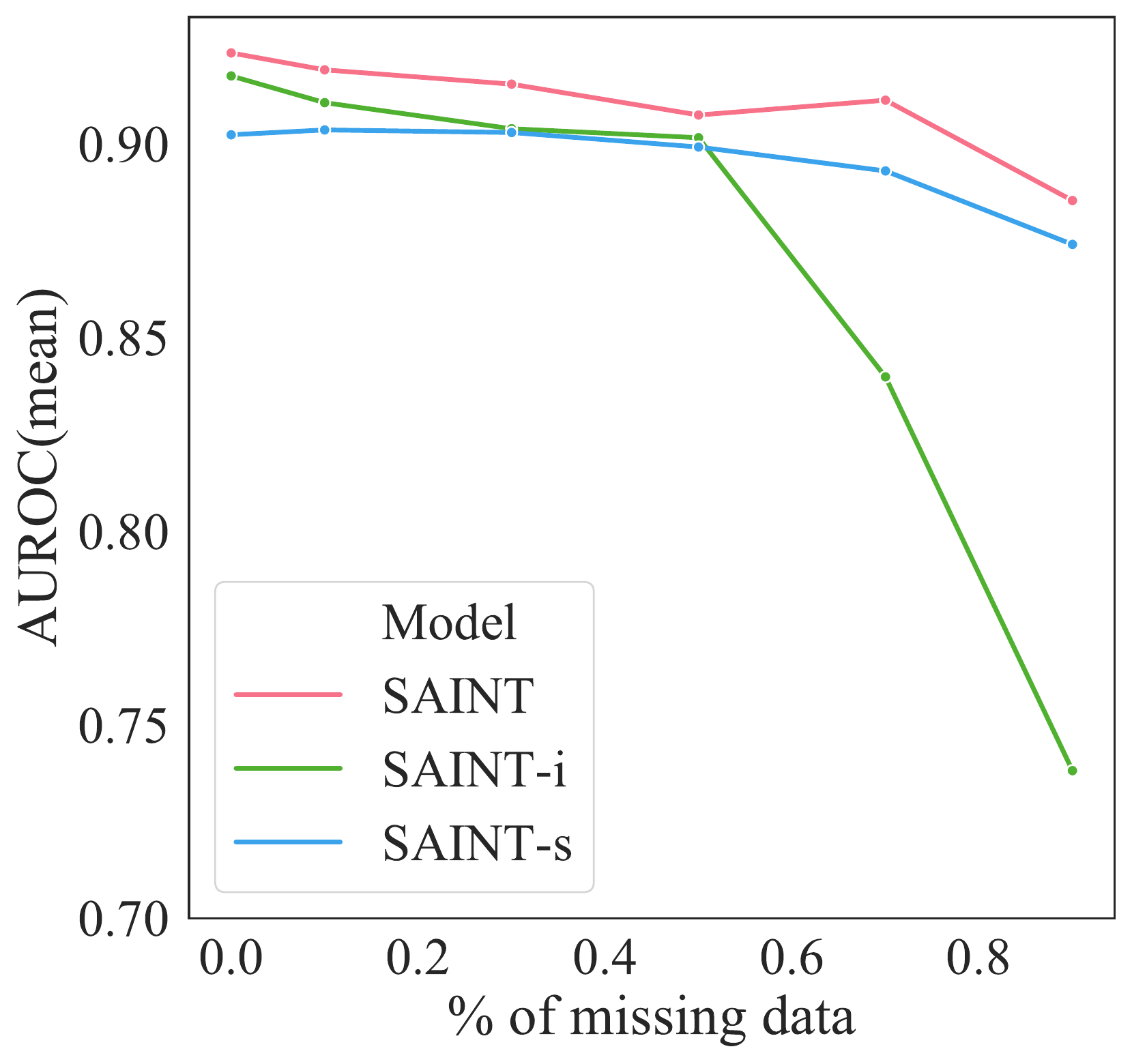}
         \caption{Trendlines of mean AUROC of SAINT's variants with varying \% of missing data.}
         \label{fig:missing_data_plot}
     \end{subfigure}
     \hfill
     \begin{subfigure}[b]{0.48\textwidth}
         \centering
         \includegraphics[width=\textwidth]{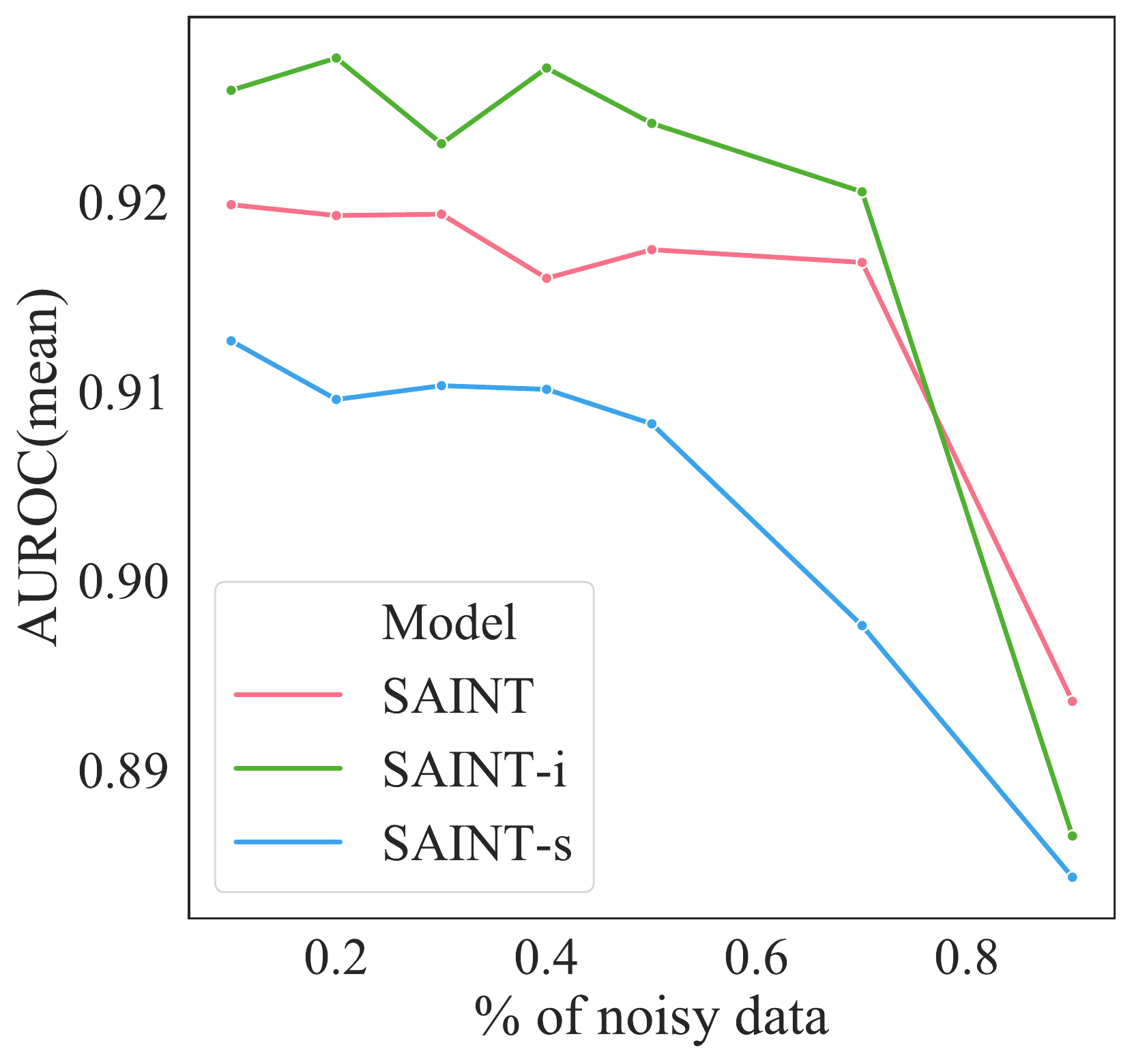}
         \caption{Trendlines of mean AUROC of SAINT's variants with varying \% of noisy data.}
         \label{fig:noisy_data_plot}
     \end{subfigure}
     \caption{Robustness of SAINT's variants to data corruptions.}
     \label{fig:SAINT_robustness}
\end{figure}

\paragraph{Corrupted training data} We show in Figure~\ref{fig:SAINT_robustness} how the mean AUROC varies as we vary the percentage of the training data that is corrupted. We consider 2 types of corruptions - missing data as shown in Figure~\ref{fig:missing_data_plot} and noisy data as shown in Figure~\ref{fig:noisy_data_plot}. We observe that the SAINT model is quite robust across both variants, and the drop in performance is minimal until 70\% of training data is corrupted. We also observe that the self-attention variant SAINT-s is more robust in the case of missing data, while the intersample attention variant SAINT-i is more robust in case of noisy data. 

\paragraph{Effect of batch size on intersample attention performance (cont.)}

\begin{figure}[h]
  \centering
  \includegraphics[width=\textwidth]{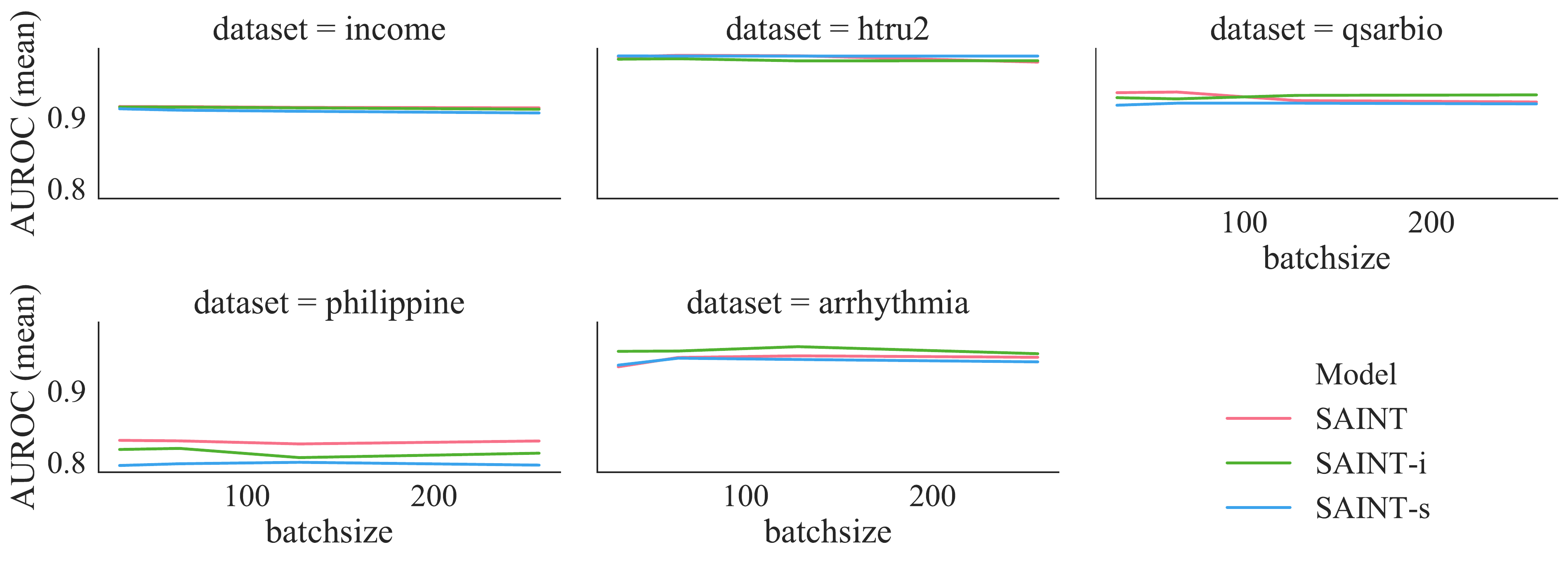}
  \caption{Trend lines of AUROC with varying training batch size. Results shown for 5 datasets}
  \label{fig:batchsize_ablation}
\end{figure}

As discussed in the main body, we examine the affect of batch size on different SAINT variants in Figure ~\ref{fig:batchsize_ablation}. We pick 5 datasets with varying numbers of features and samples. In all cases, we see that the variance in AUROC is minimal when varying the batch size from 32 to 256.
\subsection{Pre-training Ablations}

\begin{figure}[h]
  \centering
  \includegraphics[width=\textwidth]{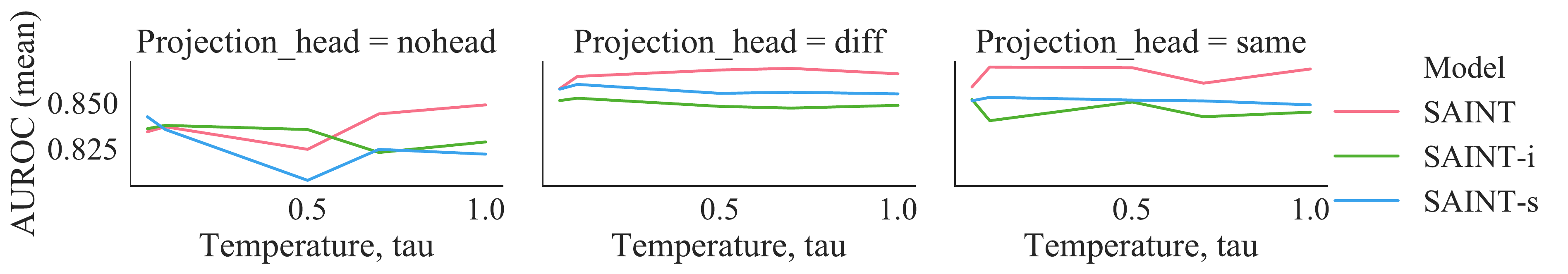}
  \caption{Temperature and Projection head ablation}
  \label{fig:temp_ablation}
\end{figure}

In Table~\ref{table:pt_ablation}, we study various configurations of pre-training components. We perform 3 primary studies: we vary (1) projection head, (2) pre-training loss, and (3) data augmentation method. Note, the final result in all 3 studies refers to the same experiment (hence the row is repeated), which is the final chosen configuration for our model. In addition to the table, in Figure ~\ref{fig:temp_ablation}, we study the connection between the temperature $\tau$ and the type of projection head.
\begin{table}[h!]
\centering
\captionsetup{width=0.7\textwidth,font=small}
\captionof{table}{Ablation studies on the pre-training pipeline of SAINT. We break down the effect of the projection head, pre-training loss, and augmentation method. We report average AUC (in \%) over 14 datasets for the case where only 50 points in the dataset are labeled. }
\resizebox{0.7\textwidth}{!}{

\begin{tabular}{c|c|ccc}
\toprule
Study              & Variation                & SAINT-s     & SAINT-i     & SAINT\\
 \toprule
\multirow{3}{*}{1} & no proj. head            & 84.26 & 83.56 & 84.90   \\
                   & weight sharing head & 85.31 & 85.20 & 86.89   \\
                   & w. diff proj. head & \textbf{86.02} & \textbf{85.26} & \textbf{86.96}   \\
\midrule
\multirow{5}{*}{2} & no pre-training          & 85.14 & 83.93 & 85.78   \\
                   & contrastive             & 85.40 & 84.42 & 85.58   \\
                   & denoising               & 84.74 & 84.93 & 86.21   \\
                   & cosine similarity               & 85.03 & 84.35 & 85.70   \\
                   & contra. + denois.  & \textbf{86.02} & \textbf{85.26} & \textbf{86.96}   \\
\midrule
\multirow{3}{*}{3} & CutMix                  & 82.80 & 84.61 & 85.37  \\ 
                   & mixup                   & 86.01 & 84.41 & 86.45   \\
                   & CutMix + mixup          & \textbf{86.02} & \textbf{85.26} & \textbf{86.96}   \\
\bottomrule
\end{tabular}
}

\label{table:pt_ablation}
\end{table}

\textbf{Effect of projection heads:} As described in Section \ref{sec:pretraining}, we use two different projection heads, $g_1(\cdot)$ and $g_2(\cdot)$, to project the contextual representations to lower dimensions and then compute contrastive losses. We study three different options for the heads: (1) distinct projection heads (2) heads with weight sharing, and 
(3) no projection heads at all. Table~\ref{table:pt_ablation} shows that using distinct projection heads performs best.

\textbf{Varying pre-training loss:} We train SAINT's variants with different loss functions, as shown in Study 2 of Table~\ref{table:pt_ablation}. We try denoising and contrastive losses, in addition to a cosine similarity loss on positive pairs (inspired by \cite{grill2020bootstrap, chen2020exploring}). The combination of contrastive and denoising consistently yields the best results in all SAINT variants.

\textbf{Varying the pre-training augmentations:} We also try to understand how important it is to use CutMix and mixup to generate augmented embeddings in the pre-training pipeline. We tinker with various configurations in Study 3 of Table~\ref{table:pt_ablation}, and we observe that using these two augmentations in unison results in the best performance across all SAINT variants.

\section{Additional interpretability plots}
\label{appendix_sec:additional_interpret}

\begin{figure}[h]
     \centering
     \begin{subfigure}[b]{0.3\textwidth}
         \centering
         ~\vspace{4mm}
         \includegraphics[width=\textwidth,valign=t]{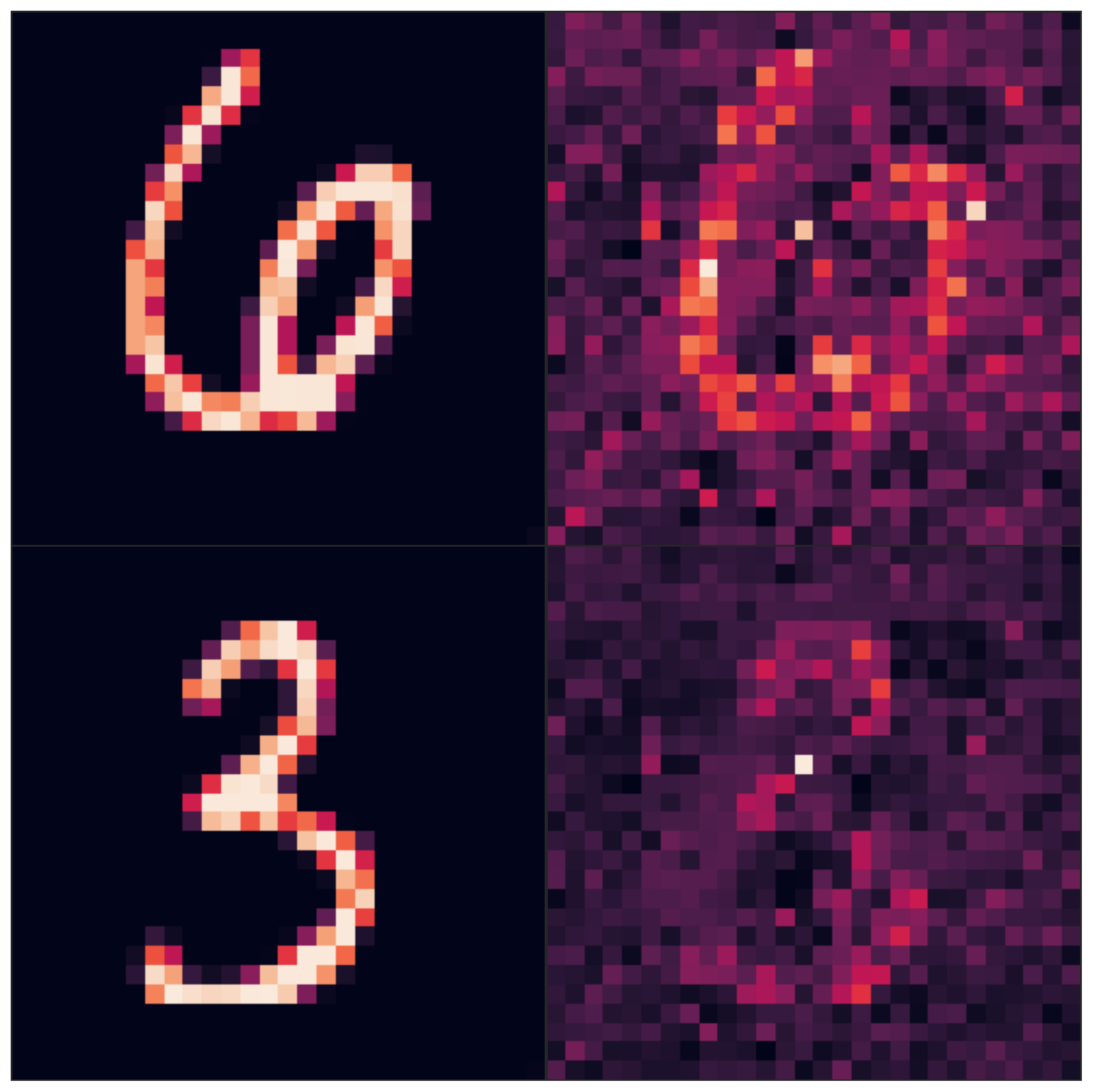}
         \caption{Self-attn. in 1 layered SAINT-s on MNIST dataset}
         \label{fig:SA_in_SAINTs_mnist}
     \end{subfigure}
     \hfill
     \begin{subfigure}[b]{0.32\textwidth}
         \centering
         \includegraphics[width=\textwidth,valign=t]{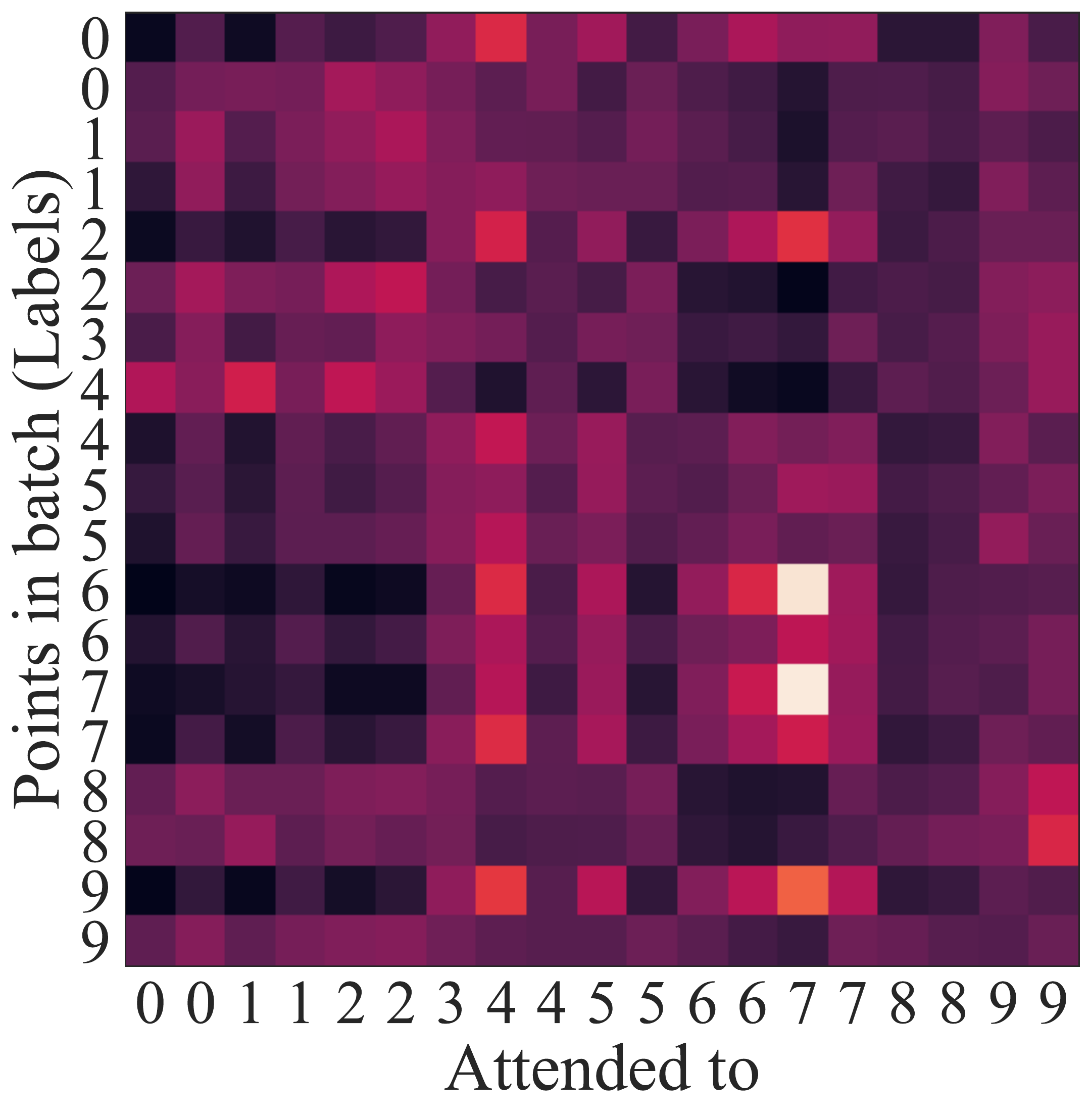}
         \caption{Intersample attn. in SAINT in Volkert dataset}
         \label{fig:ISA_in_SAINT_volkert}
     \end{subfigure}
     \hfill
     \begin{subfigure}[b]{0.32\textwidth}
         \centering
         \includegraphics[width=\textwidth,valign=t]{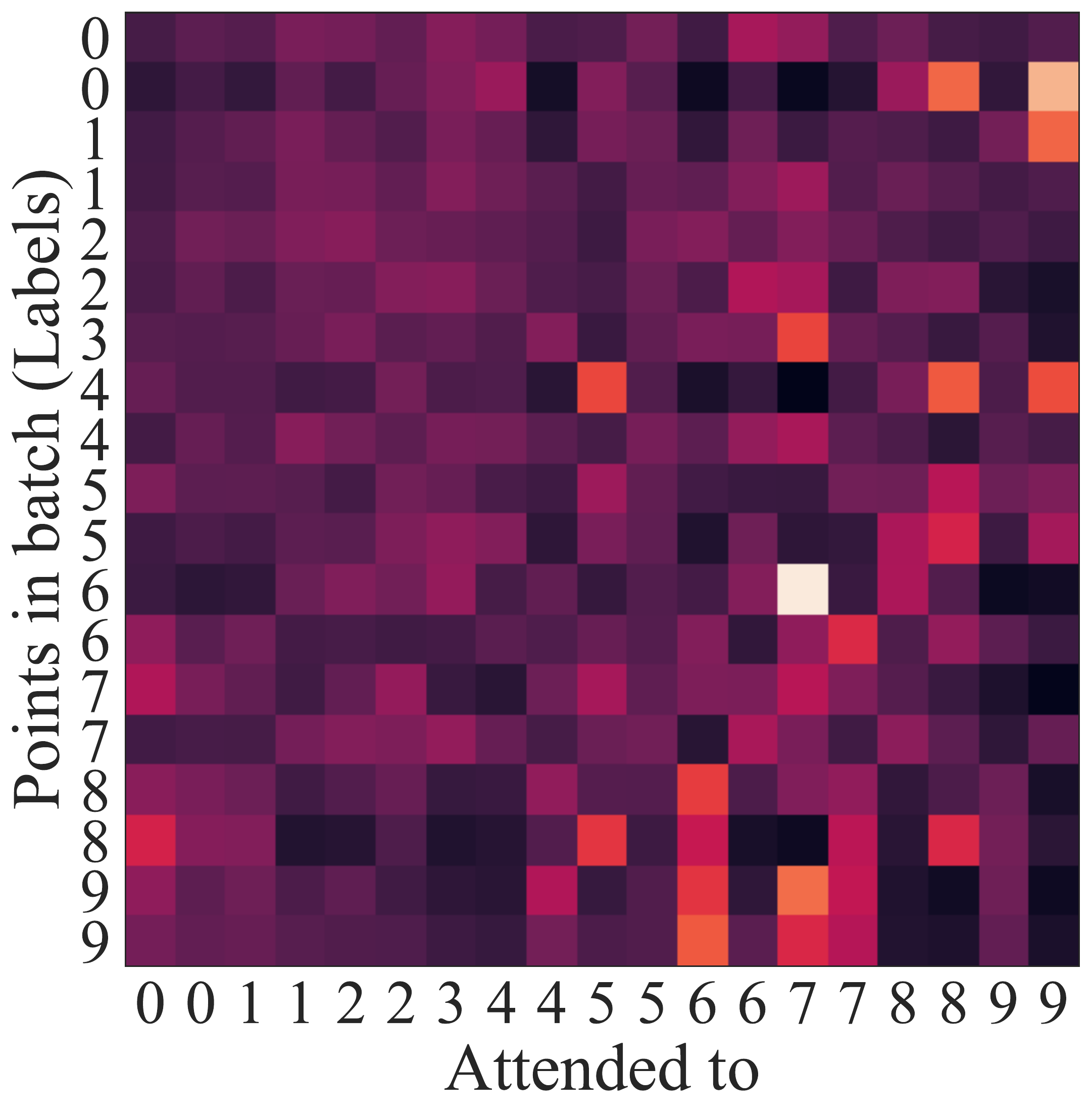}
         \caption{Intersample attn. in SAINT-i in Volkert dataset} 
         \label{fig:ISA_in_SAINTi_volkert}
     \end{subfigure}
     \captionsetup{font=small}
        \caption{Visual representations of various attention mechanisms. (a) Self-attention in SAINT-s on MNIST (b,c) Intersample attention in SAINT and SAINT-i on the Volkert dataset. }
        \label{fig:appendix_attention_plots}
\end{figure}

In Figure~\ref{fig:SA_in_SAINTs_mnist}, we show a self-attention plot for the SAINT-s variant (with $L=1$) on MNIST. The self-attention in one stage SAINT-s model behaves similar to a one stage SAINT model. However, when there are more stages, the attention in the last stage is not quite as interpretible.

In Figure~\ref{fig:ISA_in_SAINT_volkert}, we show the intersample attention between a batch of points from different classes in SAINT model on the Volkert dataset. Similarly in Figure~\ref{fig:ISA_in_SAINTi_volkert}, we show intersample attention in the SAINT-i variant on the same batch of points from the Volkert dataset. As mentioned in the main body, the intersample behaviour is not quite as sparse as that of MNIST. We hypothesize that the sparsity of the intersample attention layer depends on how separable the classes in the dataset are. (Volkert is a harder dataset than MNIST).

\begin{figure}[h]
    \centering
     \begin{subfigure}[b]{0.45\textwidth}
         \centering
         \includegraphics[width=\textwidth]{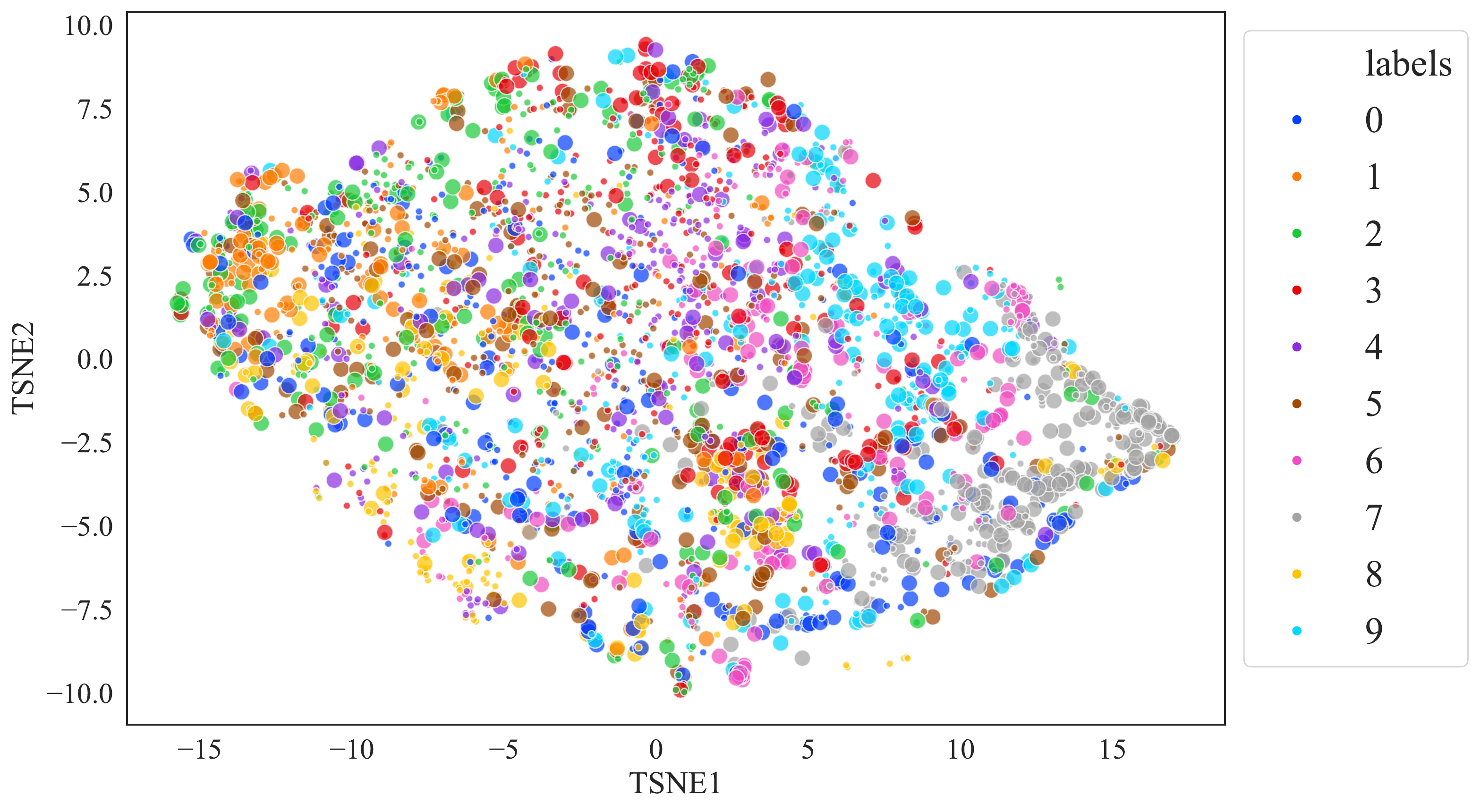}
         \label{fig:saint_tsne_volkert}
     \end{subfigure}
     \begin{subfigure}[b]{0.45\textwidth}
         \centering
         \includegraphics[width=\textwidth]{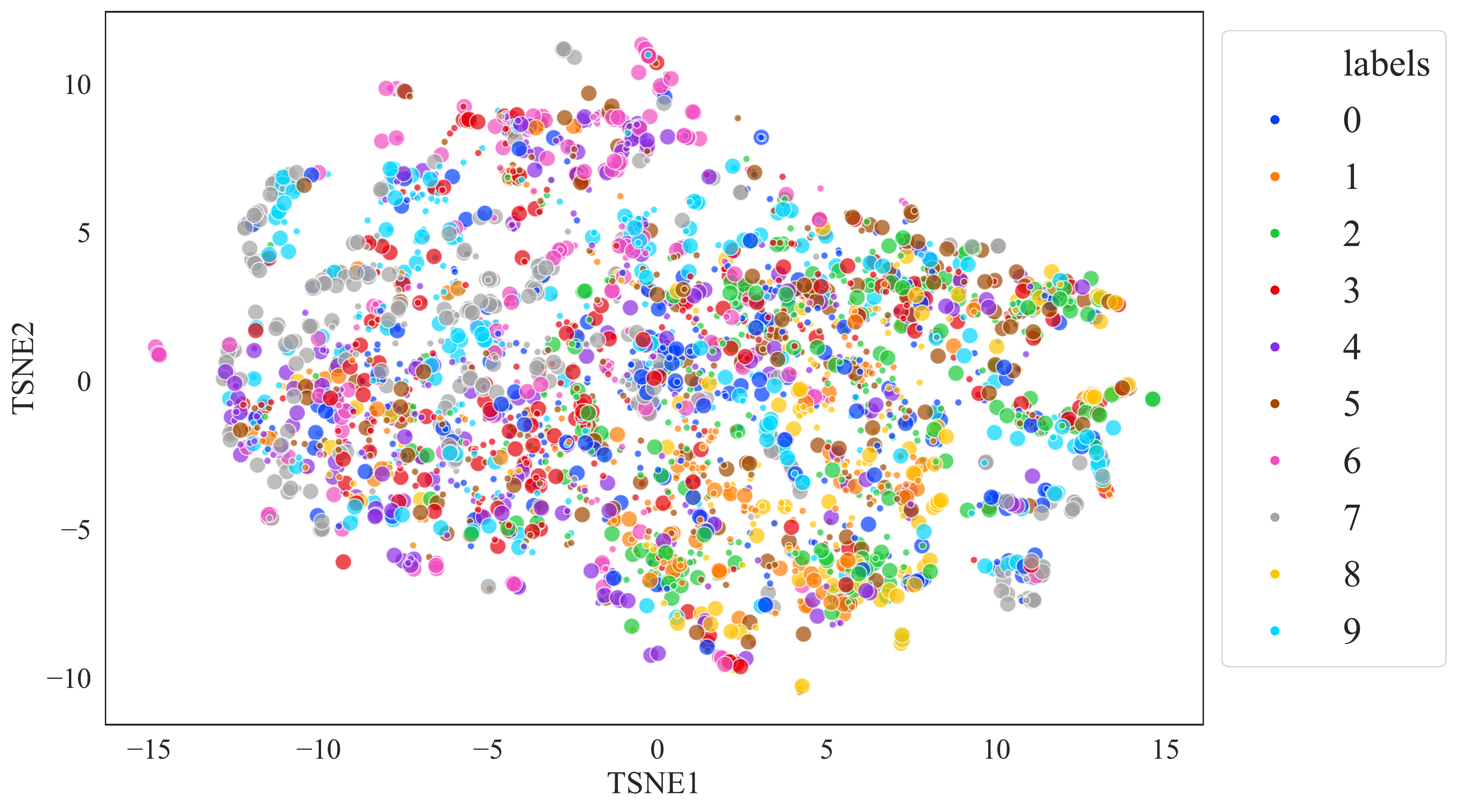}
         \label{fig:saint_i_tsne_volkert}
     \end{subfigure}
     \captionsetup{font=small}
     \vspace{-5mm}
    \caption{A t-SNE visualization of \textit{value} vectors in intersample attention layers of SAINT~(left) and SAINT-i~(right) on the Volkert dataset. We plot 3000 points in each figure, with classes uniformly represented. Unlike MNIST, all classes are uniformly attended to in this dataset.}
    \label{fig:attention_tsne_volkert}
\end{figure}

In Figure~\ref{fig:attention_tsne_volkert}, we show the t-SNE plots on value vectors for SAINT and SAINT-i variants on Volkert. Unlike MNIST, all the classes are attended to equally.

\end{document}